%% file: egpaper_for_review.tex
\documentclass[10pt,twocolumn,letterpaper]{article}

\usepackage{iccv}
\usepackage{times}
\usepackage{epsfig}
\usepackage{graphicx}
\usepackage{amsmath}
\usepackage{amssymb}
\usepackage{url}            
\usepackage{booktabs}       
\usepackage{amsfonts}       
\usepackage{nicefrac}       
\usepackage{microtype}      
\usepackage{amsmath}
\usepackage{bm}
\usepackage{bbold}
\usepackage{subfig}
\usepackage{rotating}
\usepackage[export]{adjustbox}
\usepackage{color}
\usepackage[dvipsnames]{xcolor}
\newcommand{\ra}[1]{\renewcommand{\arraystretch}{#1}}

\makeatletter
\let\@fnsymbol\@arabic
\makeatother

\newif\ifkeepComments
\keepCommentstrue

\usepackage[pagebackref=true,breaklinks=true,letterpaper=true,colorlinks,bookmarks=false]{hyperref}

\iccvfinalcopy 

\def\iccvPaperID{2531} 

\ificcvfinal\fi
\begin{document}

\title{Detecting Unseen Visual Relations Using Analogies}

\author{Julia Peyre\footnotemark[1] \textsuperscript{,}\footnotemark[2]
 \qquad Ivan Laptev\footnotemark[1] \textsuperscript{,}\footnotemark[2] \qquad Cordelia Schmid\footnotemark[2] \textsuperscript{,}\footnotemark[4] \qquad Josef Sivic\footnotemark[1] \textsuperscript{,}\footnotemark[2] \textsuperscript{,}\footnotemark[3] \\
}

\maketitle

\footnotetext[1]{D\'epartement d'informatique de l'ENS, Ecole normale sup\'erieure, CNRS, PSL Research University, 75005 Paris, France.}
\footnotetext[2]{INRIA}
\footnotetext[3]{Czech Institute of Informatics, Robotics and Cybernetics at the Czech Technical University in Prague.}
\footnotetext[4]{Univ. Grenoble Alpes, Inria, CNRS, Grenoble INP, LJK, 38000 Grenoble, France.}

\begin{abstract}
\input{abstract}

\end{abstract}

\begin{figure}
\includegraphics[trim={0cm 0cm 0cm 0cm},clip,width=\linewidth]{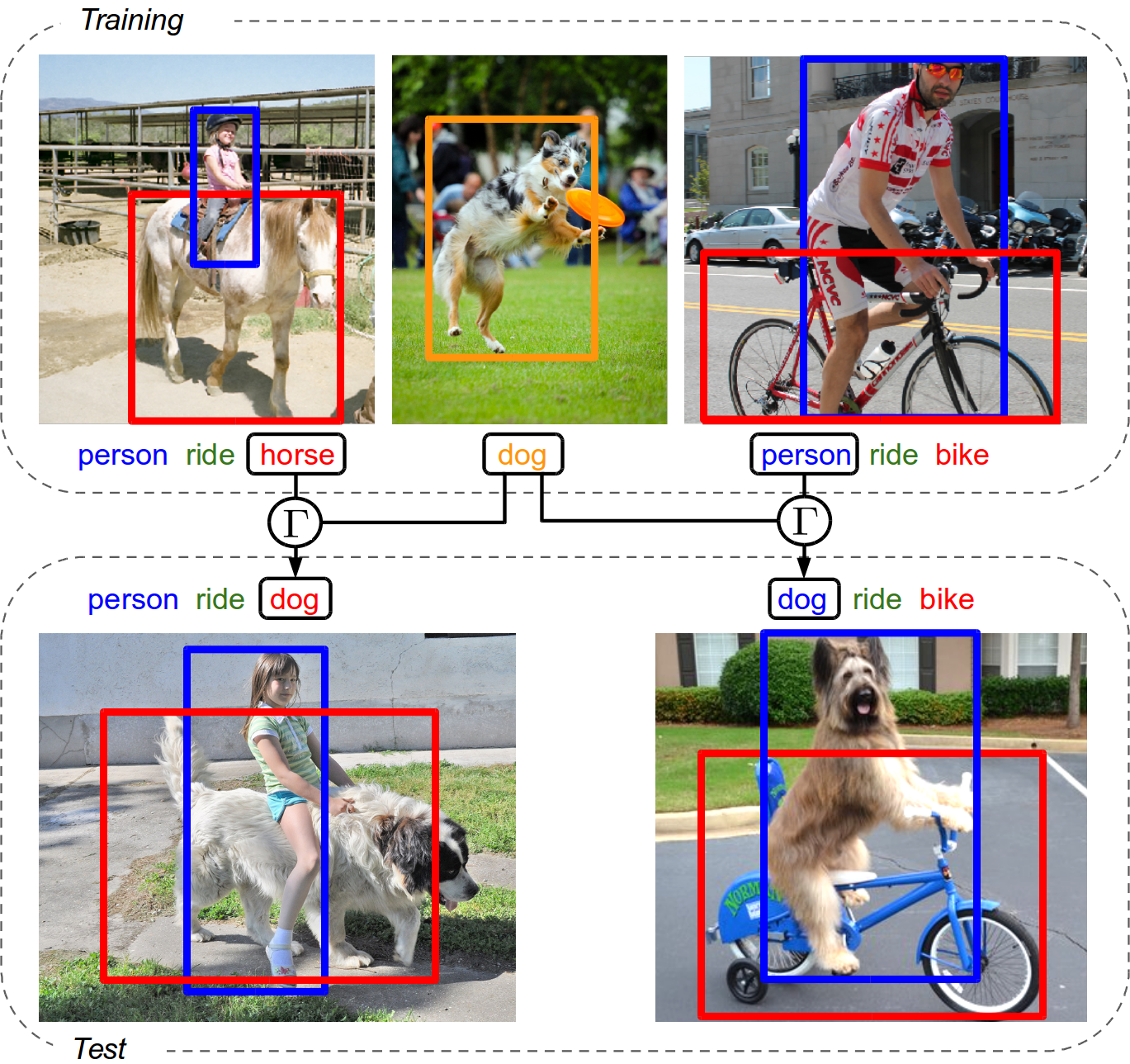}
\vspace{-0.6cm}
\caption{\small Illustration of transfer by analogy with our model described in~\ref{part:model_2}. We transfer visual representations of relations seen in the training set such as ``person ride horse" to represent new unseen relations in the test set such as ``person ride dog".  }
\label{fig:teaser}
\vspace{-.45cm}
\end{figure}

\vspace{-2ex}
\section{Introduction}
\input{introduction}

\begin{figure*}[t]
\vspace*{-5mm}
\includegraphics[trim={0.5cm 4.5cm 1.5cm 0cm},clip,width=\linewidth]{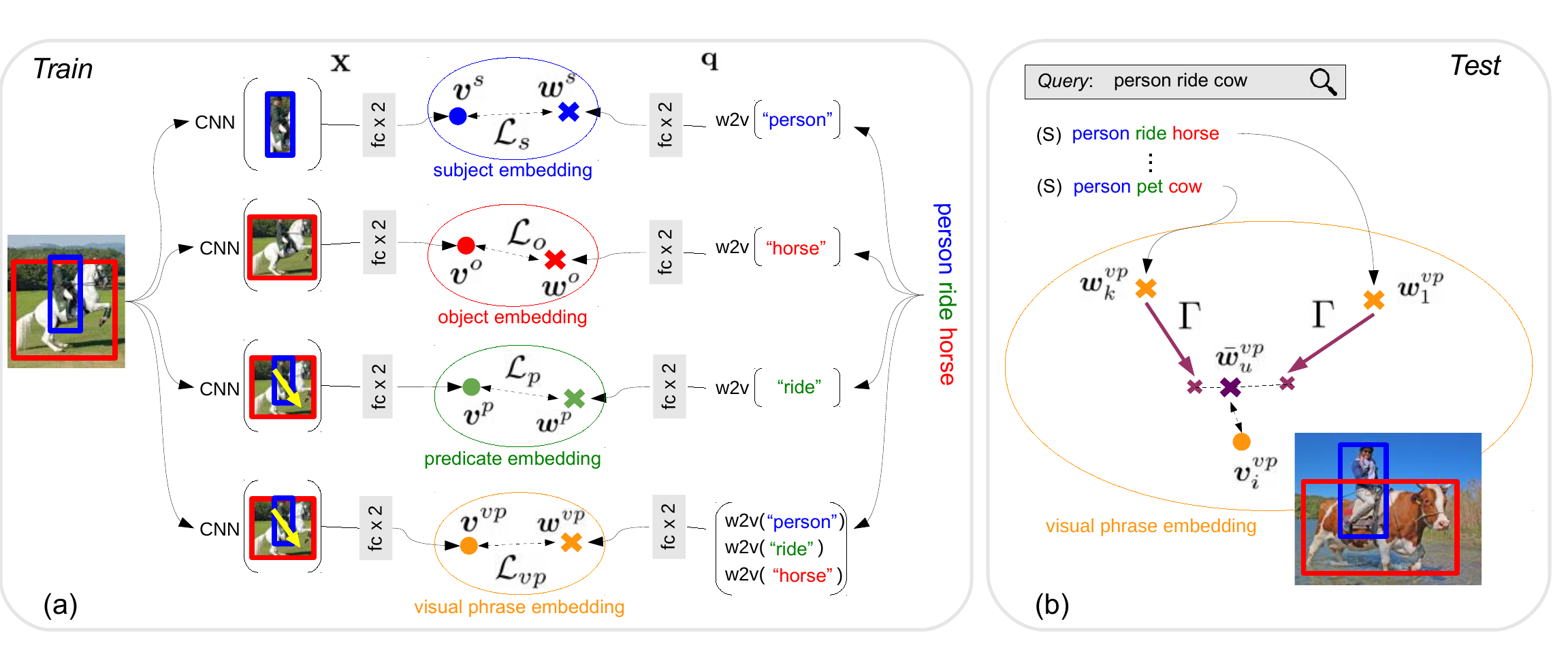}
\vspace*{-9.5mm}
\caption{\small
{\bf Model overview.} Our model consists of two parts : (a) learning embedding  spaces for subject, object, predicate and visual phrase by optimizing the joint loss $\mathcal{L}_{joint} = \mathcal{L}_s + \mathcal{L}_o + \mathcal{L}_p + \mathcal{L}_{vp}$ combining the input visual $\mathbf{x}$ and language $\mathbf{q}$ representations; (b) at test time, we are given a new unseen triplet (``person ride cow"). We find similar but seen triplets (``person ride horse" and ``person pet cow"), transform their embeddings $\bm{w}_k^{vp}$ with analogy transformation $\Gamma$ to compute an estimate of the embedding $\bar{\bm{w}}_u^{vp}$ of the triplet ``person ride cow" and use this estimated embedding to retrieve relevant images by nearest neighbour search on the (embedded) visual descriptors $v_i^{vp}$.}
\label{fig:model}
\vspace*{-4mm}
\end{figure*}

\section{Related work}
\input{related_work}


\section{Model}
\label{part:model}
\input{model}

\input{experiments_fig_hico}

\section{Experiments}
\label{part:experiments}
\input{experiments}

\section{Conclusion}
\input{conclusion}


\vspace{-3ex}
\paragraph{Acknowledgements.}
{\small{This work was partly supported by ERC grants Activia (no.307574),
LEAP (no.336845), Allegro (no.320559), CIFAR Learning in Machines \& Brains program, the MSR-Inria joint lab, Louis Vuitton ENS Chair on Artificial Intelligence, DGA project DRAAF, and European Regional Development Fund under the project IMPACT (reg. no. CZ.02.1.01/0.0/0.0/15\_003/0000468)}}

{\small
\bibliographystyle{ieee_fullname}
\bibliography{refs}
}

\clearpage
\appendix
\input{appendix}

\end{document}

%% file: abstract.tex
We seek to detect visual relations in images of the form of triplets $t=~(subject,~predicate,~object)$, such as ``person riding dog", where training examples of the individual entities are available but their combinations are unseen at training. This is an important set-up due to the combinatorial nature of visual relations : collecting sufficient training data for all possible triplets would be very hard. 
The contributions of this work are three-fold. First, we learn a representation of visual relations that combines (i) individual embeddings for subject, object and predicate together with (ii) a visual phrase embedding that represents the relation triplet. 
Second, we learn how to transfer visual phrase embeddings from existing training triplets to unseen test triplets using analogies between relations that involve similar objects. 
Third, we demonstrate the benefits of our approach on three challenging datasets : on HICO-DET, our model achieves significant improvement over a strong baseline for both frequent and unseen triplets, and we observe similar improvement for the retrieval of unseen triplets with out-of-vocabulary predicates on the COCO-a dataset as well as the challenging unusual triplets in the UnRel dataset.

%% file: introduction.tex
Understanding interactions between objects is one of the fundamental problems in visual recognition. To retrieve images given a complex language query such as ``a woman sitting on top of a pile of books" we need to recognize individual entities ``woman" and ``a pile of books" in the scene, as well as understand what it means to ``sit on top of something". In this work we aim to recognize and localize unseen interactions in images, as shown in Figure~\ref{fig:teaser}, where the individual entities (``person", ``dog", ``ride") are available at training, but not in this specific combination. Such ability is important in practice given the combinatorial nature of visual relations where we are unlikely to obtain sufficient training data for all possible relation triplets. 

Existing methods~\cite{Dai17,Yikang17bis,Lu16} to detect visual relations in the form of triplets $t=(subject,predicate,object)$ typically learn generic detectors for each of the entities, i.e.\ a separate detector is learnt for subject (e.g. ``person"), object (e.g. ``horse") and predicate (e.g. ``ride"). The outputs of the individual detectors are then aggregated at test time.  This {\em compositional approach} can detect unseen triplets, where subject, predicate and object are observed separately but not in the specific combination. However, it often fails in practice~\cite{Peyre17,Zhang17}, due to the large variability in appearance of the visual interaction that often heavily depends on the objects involved; it is indeed difficult for a single ``ride" detector to capture visually different relations such as ``person ride horse" and ``person ride bus". 

An alternative approach~\cite{Sadeghi2011} is to treat the whole triplet as a single entity, called a visual phrase, and learn a separate detector for each of the visual phrases. For instance, separate detectors would be learnt for relations ``person ride horse" and ``person ride surfboard". While this approach better handles the large variability of visual relations, it requires training data for each triplet, which is hard to obtain as visual relations are combinatorial in their nature and many relations are unseen in the real world. 

In this work we address these two key limitations. First, what is the right representation of visual relations to handle the large variability in their appearance, which depends on the entities involved? Second, how can we handle the scarcity of training data for unseen visual relation triplets?
To address the first challenge, we develop a hybrid model that combines compositional and visual phrase representations. 
More precisely, we learn a compositional representation for subject, object and predicate by learning separate visual-language embedding spaces where each of these entities is mapped close to the language embedding of its associated annotation. In addition, we also learn a relation triplet embedding space where visual phrase representations are mapped close to the language embedding of their corresponding triplet annotations. At test time, we aggregate outputs of both compositional and visual phrase models. 
To address the second challenge, we learn how to transfer visual phrase embeddings from existing training triplets to unseen test triplets using analogies between relations that involve similar objects.
For instance, as shown in Figure~\ref{fig:teaser}, we recognize the unseen triplet ``person ride dog" by using the visual phrase embedding for triplet ``person ride horse" after a transformation that depends on the object embedding for ``dog" and ``horse".
Because we transfer training data only from triplets that are visually similar, we expect transferred visual phrase detectors to better represent the target relations compared to a generic detector for a relation ``ride" that may involve also examples of ``person ride train" and ``person ride surfboard".  

\vspace{-.4cm}
\paragraph{Contributions.} Our contributions are three fold. First, we take advantage of both the compositional and visual phrase representations by learning complementary visual-language embeddings for subject, object, predicate and the visual phrase. Second, we develop a model for transfer by analogy to obtain visual-phrase embeddings of never seen before relations. Third, we perform experimental evaluation on three challenging datasets where we demonstrate the benefits of our approach on both frequent and unseen relations. 

%% file: related_work.tex
\paragraph{Visual relation detection.}
Learning visual relations belongs to a general class of problems on relational reasoning~\cite{Bansal2017,Battaglia2016,jenatton2012,Kipf2016,Santoro17} that aim to understand how entities interact.
In the more specific set-up of visual relation detection, the approaches can be divided into two main groups: (i) compositional models, which learn detectors for subject, object and predicates separately and aggregate their outputs; (ii) and visual phrase models, which learn a separate detector for each visual relation. 
Visual phrase models such as~\cite{Sadeghi2011} have demonstrated better robustness to the visual diversity of relations than compositional models. However, with the introduction of datasets with a larger vocabulary of objects and predicates~\cite{Chao15,Krishna2016}, visual phrase approaches have been facing severe difficulties as most relations have very few training examples. Compositional methods~\cite{iCAN,Gkioxari18,Johnson15a,Lu16,Peyre17,Qi18,Shen18}, which allow sharing knowledge across triplets, have scaled better but do not cope well with unseen relations. 
To increase the expressiveness of the generic compositional detectors, recent works have developed models of statistical dependencies between the subject, object and predicate, using, for example, graphical models~\cite{Dai17,Yikang17bis}, language distillation~\cite{Yu17}, or semantic context~\cite{Bohan17}. 
Others~\cite{Atzmon16,Divvala2014,plummerPLCLC2016,sadeghi2015viske} have proposed to combine unigram detectors with higher-order composites such as bigrams (subject-predicate, predicate-object).
In contrast to the above methods that model a discrete vocabulary of labels, we learn visual-semantic (language) embeddings able to scale to out-of-vocabulary relations and to benefit from powerful pre-learnt language models.

\vspace{-.37cm}
\paragraph{Visual-semantic embeddings.} Visual-semantic embeddings have been successfully used for image captioning and retrieval \cite{Karpathy2014,Karpathy2014a}. With the introduction of datasets annotated at the region level~\cite{Krishna2016,Plummer15}, similar models have been applied to align image regions to fragments of sentences~\cite{Izadinia2015,Wang16}.
In contrast, learning embeddings for visual relations still remains largely an open research problem with recent work exploring, for example, relation representations using deformations between subject and object embeddings~\cite{Zhang17}. 
Our work is, in particular, related to models~\cite{Ji18} learning separate visual-semantic spaces for subject, object and predicate. However, in contrast to~\cite{Ji18}, we additionally learn a visual phrase embedding space to better deal with appearance variation of visual relations, and develop a model for analogy reasoning to infer embeddings of unseen triplets. 

\vspace{-.37cm}
\paragraph{Unseen relations and transfer learning.} 
Learning visual phrase embeddings suffers from the problem of lack of training data for unseen relations. This has been addressed by learning factorized object and predicate representations~\cite{Hwang18} or by composing classifiers for relations from simpler concepts~\cite{Kato18,Misra17}. 
In contrast, our approach transfers visual relation representations from seen examples to unseen ones in a similar spirit to how previous work dealt with inferring classifiers for rare objects~\cite{Aytar11}. The idea of sharing knowledge from seen to unseen triplets to compensate for the scarcity of training data has been also addressed in~\cite{ramanathan15} by imposing constraints on embeddings of actions. 
Different from this work, we formulate the transfer as an analogy between relation triplets. To achieve that, we build on the computational model of analogies developed in~\cite{Reed2015a} but extend it to representations of visual relations. 
This is related to~\cite{Sadeghi2015} who also learn visual analogies as vector operations in an embedding space, but only consider visual inputs while we learn analogy models for joint image-language embeddings.

%% file: model.tex
In this section we describe our model for recognizing and localizing visual relations in images. As illustrated in Figure~\ref{fig:model}, our model consists of two parts. First, we learn different visual-language embedding spaces for the subject ($s$), the object ($o$), the predicate ($p$) and the visual phrase ($vp$), as shown in Figure~\ref{fig:model}(a). We explain how to train these embeddings in Section~\ref{part:model_1}. Second, we transfer visual phrase embeddings of seen triplets to unseen ones with analogy transformations, as shown in Figure~\ref{fig:model}(b). In Section~\ref{part:model_2} we explain how to train the analogy transformations and form visual phrase embeddings of new unseen triplets at test time. 

\vspace{-2.5ex}
\paragraph{Notation for relation triplets.}
The training dataset consists of $N$ candidate pairs of bounding boxes, each formed by a subject candidate bounding box proposal and object candidate bounding box proposal. Let $\mathcal{V}_s$, $\mathcal{V}_o$ and $\mathcal{V}_p$ be the vocabulary of subjects, objects and predicates, respectively. We call $\mathcal{V}_{vp} = \mathcal{V}_s \times \mathcal{V}_p \times \mathcal{V}_o$ the vocabulary of triplets. A triplet $t$ is of the form $t=(s,p,o)$, e.g. $t = (person, ride, horse)$. Each pair of candidate subject and object bounding boxes, $i \in \{1,...,N\}$, is labeled by a vector $(y_t^i)_{t \in \mathcal{V}_{vp}}$ where $y_t^i=1$ if the $i^{th}$ pair of boxes could be described by relation triplet $t$, otherwise $y_t^i=0$. The labels for subject, object and predicate naturally derive from the triplet label. 

\subsection{Learning representations of visual relations}
\label{part:model_1}

We represent visual relations in joint visual-semantic embedding spaces at different levels of granularity : (i) at the unigram level, where we use separate subject, object and predicate embeddings, and (ii) at the trigram level using an a visual phrase embedding of the whole triplet. Combining the different types of embeddings results in a more powerful representation of visual relations as will be shown in section~\ref{part:experiments}. 
In detail, as shown in Figure~\ref{fig:model}(a), the input to visual embedding functions (left) is a candidate pair of objects $i$ encoded by its visual representation $\bm{\mathbf{x}}_i \in \mathbb{R}^{d_v}$. This representation is built from (i) pre-computed appearance features obtained from a CNN trained for object detection and (ii) a representation of the relative spatial configuration of the object candidates. The language embeddings (right in Figure~\ref{fig:model}(a)) take as input a triplet $t$ encoded by its language representation $\bm{\mathbf{q}}_t \in \mathbb{R}^{d_q}$ obtained from pre-trained word embeddings. We provide more details about these representations in~\ref{part:implementation_details}. Next we give details of the embedding functions. 

\vspace{-2.5ex}
\paragraph{Embedding functions.}
Our network projects the visual features $\bm{\mathbf{x}}_i$ and language features $\bm{\mathbf{q}}_t$ into separate spaces for the subject ($s$), the object ($o$), the predicate ($p$) and the visual phrase ($vp$). For each input type $b \in \{ s,o,p,vp \}$, we embed the visual features and language features into a common space of dimensionality $d$ using projection functions
\begin{align}
\label{eq:proj}
\bm{v}_i^b = f^b_v(\bm{\mathbf{x}}_i),  \\
\label{eq:proj2}
\bm{w}_t^b = f^b_w(\bm{\mathbf{q}}_t),
\end{align}
where $\bm{v}_i^b$ and $\bm{w}_t^b$ are the output visual and language representations, and the projection functions $f^b_v : \mathbb{R}^{d_v} \rightarrow \mathbb{R}^{d}$ and $f^b_w : \mathbb{R}^{d_q} \rightarrow \mathbb{R}^{d}$ are 2-layer perceptrons, with ReLU non linearities and Dropout, inspired by~\cite{Wang16}.  
Additionally, we L2 normalize the output language features while the output visual features are not normalized, which we found to work well in practice. 
 
\vspace{-2.5ex}
\paragraph{Training loss.}
We train parameters of the embedding functions $(f^b_v, f^b_w)$ for each type of input $b$ (i.e subject, object, predicate and visual phrase) by maximizing log-likelihood 
\begin{align}
\label{eq:loss}
\mathcal{L}_b =& \sum_{i=1}^N \sum_{t \in \mathcal{V}_b} \mathbb{1}_{y_t^i=1} \log \left( \frac{1}{1+e^{-{\bm{w}_t^{b}}^{T} \bm{v}_i^b}} \right) \notag \\
 +& \sum_{i=1}^N \sum_{t \in \mathcal{V}_b} \mathbb{1}_{y_t^i=0} \log \left( \frac{1}{1+e^{{\bm{w}_t^b}^T \bm{v}_i^b}} \right),
\end{align}
where the first attraction term pushes closer visual representation $\bm{v}_i^b$ to its correct language representation $\bm{w}_t^b$ and the  
 second repulsive term pushes apart visual-language pairs that do not match. As illustrated in Figure~\ref{fig:model}, we have one such loss for each input type and optimize the joint loss that sums the individual loss functions $\mathcal{L}_{joint} = \mathcal{L}_s + \mathcal{L}_o + \mathcal{L}_p + \mathcal{L}_{vp}$. A similar loss function has been used in~\cite{mikolov2013distributed} to learn word representations, while visual-semantic embedding models~\cite{Karpathy2014a,Wang16} typically use triplet ranking losses. Both loss functions work well, but we found embeddings trained with log-loss~\eqref{eq:loss} easier to combine across different input types as their outputs are better calibrated. 

\vspace{-2.8ex}
\paragraph{Inference.}
At test time, we have a language query in the form of triplet $t$ that we embed as $(\bm{w}_t^b)_b$ using Eq.~\eqref{eq:proj2}. Similarly, pairs $i$ of candidate object boxes in the test images are embedded as $(\bm{v}_i^b)_b$ with Eq.~\eqref{eq:proj}. Then we compute a similarity score $S_{t,i}$ between the triplet query $t$ and the candidate object pair $i$ by aggregating predictions over the different embedding types $b \in \{s,p,o,vp\}$ as
\begin{equation}
S_{t,i} = \prod_{b \in \{s,p,o,vp\}} \frac{1}{1+e^{-{\bm{w}_t^b}^T \bm{v}_i^{b}}}.
\end{equation}

\begin{figure}[t]
\includegraphics[trim={20cm 16cm 22cm 4cm},clip,width=\linewidth]{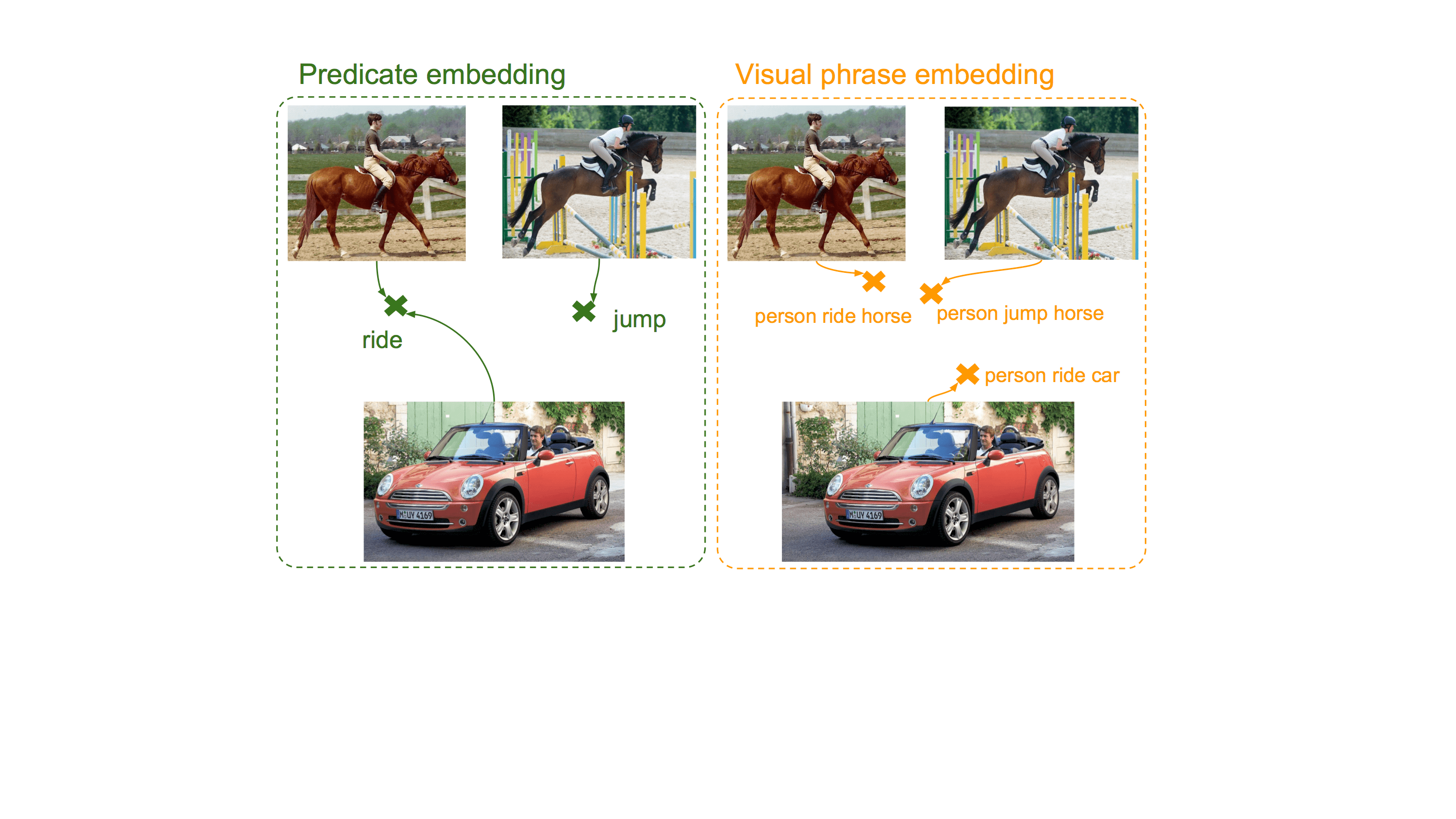}
\vspace{-5ex}
\caption{\small Illustration of the differences between predicate ($p$) (left) and visual phrase ($vp$) (right) embeddings. In the $p$ space, visually different relations such as ``person ride horse" and ``person ride car" map to the same location defined by the predicate ``ride". In contrast, they are mapped to distinct locations in the visual phrase space that considers the entire relation triplet.}
\label{fig:similarity}
\vspace{-2ex}
\end{figure}

\vspace{-3.5ex}
\paragraph{Interpretation of embedding spaces.}
The choice of learning different embedding spaces for subject, object, predicate and visual phrase is motivated by the observation that each type of embedding captures different information about the observed visual entity. 
In Figure~\ref{fig:similarity} we illustrate the advantage of learning separate predicate ($p$) and visual-phrase ($vp$) embedding spaces. In the $p$ space, visual entities corresponding to ``person ride horse" and ``person ride car" are mapped to the same point, as they share the same predicate ``ride". In contrast, in the $vp$ space, the same visual entities are mapped to two distinct points. 
This property of the $vp$ space is desirable to handle both language polysemy (i.e., ``ride" has different visual appearance depending on the objects involved and thus should not be mapped into a single point) and synonyms (i.e., ``person jump horse" and ``person ride horse" projections should be close even if they do not share the same predicate). 

\subsection{Transferring embeddings to unseen triplets by analogy transformations}
\label{part:model_2}

We propose to explicitly transfer knowledge from seen triplets at training to new unseen triplets at test time by analogy reasoning. The underlying intuition is that if we have seen examples of ``person ride horse", it might be possible to use this knowledge to recognize the relation ``person ride cow", as ``horse" and ``cow" have similar visual appearance. 
As illustrated in Figure~\ref{fig:model}(b), this is implemented as an {\em analogy transformation} in the visual phrase embedding space, where a representation of the source triplet (e.g. ``person ride horse") is transformed to form a representation of target triplet (e.g. ``person ride cow").
    There are two main steps in this process. First, we need to learn how to perform the analogy transformation of one visual phrase embedding (e.g. ``person ride horse") to another (e.g. ``person ride cow"). Second, we need to identify which visual phrases are suitable for such transfer by analogy. For example,  to form a representation of a new relation ``person ride cow" we want to transform the representation of ``person ride horse" but not ``person ride bus". We describe the two steps next. 

\vspace{-2.5ex}
\paragraph{Transfer by analogy.} To transform the visual phrase embedding $\bm{w}^{vp}_{t}$ of a source triplet $t=(s,p,o)$ to the visual phrase embedding $\bm{w}^{vp}_{t'}$ of a target triplet $t'=(s',p',o')$ we learn a transformation $\Gamma$ such that
\begin{equation}
\label{eq:correction}
\bm{w}^{vp}_{t'} = \bm{w}^{vp}_{t} + \Gamma(t,t').
\end{equation}
Here, $\Gamma$ could be interpreted as a correction term that indicates how to transform $\bm{w}^{vp}_{t}$ to $\bm{w}^{vp}_{t'}$ in the joint visual-semantic space $vp$ to compute a target relation triplet $t'$ that is analogous to source triplet $t$. This relates to neural word representations such as~\cite{mikolov2013distributed} where word embeddings of similar concepts can be linked by arithmetic operations such as $``king" - ``man" + ``woman" = ``queen"$. Here, we would like to perform operations such as $``person~ride~horse" - ``horse" + ``cow" = ``person~ride~cow"$. 

\vspace{-2.5ex}
\paragraph{Form of $\Gamma$.}
To relate the visual phrase embeddings of $t$ and $t'$ through $\Gamma$ we take advantage of the decomposition of the triplet into subject, predicate and object. 
In detail, we use the visual phrase embeddings of individual subject, predicate and object to learn how to relate the visual phrase embeddings of triplets.
Using this structure, we redefine the analogy transformation given by Eq.~\eqref{eq:correction} as
\begin{align}
\label{eq:gamma}
\bm{w}^{vp}_{t'} = \bm{w}^{vp}_{t}  +  \Gamma \begin{bmatrix}
           \bm{w}^{vp}_{s'} - \bm{w}^{vp}_{s} \\
           \bm{w}^{vp}_{p'} - \bm{w}^{vp}_{p} \\           
           \bm{w}^{vp}_{o'} - \bm{w}^{vp}_{o}
         \end{bmatrix},
\end{align}
where $t=(s,p,o)$ and $t'=(s',p',o')$ denote the source and target triplet, and $\bm{w}^{vp}_{s}$, $\bm{w}^{vp}_{p}$, $\bm{w}^{vp}_{o}$ are visual phrase embeddings of subject, predicate and object, respectively, constructed using Eq.~\eqref{eq:proj2} as $\bm{w}^{vp}_{s} = f^{vp}_w(\mathbf{q}_{[s,0,0]})$,  $\bm{w}^{vp}_{p} = f^{vp}_w(\mathbf{q}_{[0,p,0]})$, $\bm{w}^{vp}_{o} = f^{vp}_w(\mathbf{q}_{[0,0,o]})$. Here $[s,0,0]$ denotes the concatenation of word2vec embeddings of subject $s$ with two vectors of zeros of size $d$.
For example, the analogy transformation of $t=(person,ride,horse)$ to $t'=(person,ride,camel)$ using Eq.~\eqref{eq:gamma} would result in 
\begin{equation}
\bm{w}^{vp}_{t'} = \bm{w}^{vp}_{t} + \Gamma \begin{bmatrix}
           \bm{0} \\
           \bm{0} \\           
           \bm{w}^{vp}_{camel} - \bm{w}^{vp}_{horse}
         \end{bmatrix}.
\end{equation}
Intuitively, we would like $\Gamma$ to encode how the change of objects, observable through the embeddings of source and target objects, $\bm{w}^{vp}_{o}$, $\bm{w}^{vp}_{o'}$, influences the source and target triplet embeddings $\bm{w}^{vp}_{t}$, $\bm{w}^{vp}_{t'}$. Please note that here we have shown an example of a transformation resulting from a change of object, but our formulation, given by Eq.~\eqref{eq:gamma}, allows for changes of subject or predicate in a similar manner.
While different choices for $\Gamma$ are certainly possible, we opt for
\begin{equation}
\label{eq:gamma_reg}
\Gamma(t,t') = MLP \begin{bmatrix}
           \bm{w}^{vp}_{s'} - \bm{w}^{vp}_{s} \\
           \bm{w}^{vp}_{p'} - \bm{w}^{vp}_{p} \\
           \bm{w}^{vp}_{o'} - \bm{w}^{vp}_{o} 
         \end{bmatrix},
\end{equation}
where MLP is a 2-layer perceptron without bias. We also compare different forms of $\Gamma$ in Section~\ref{part:experiments}. 

\vspace{-2.5ex}
\paragraph{Which triplets to transfer from?}
We wish to apply the transformation by analogy $\Gamma$ only between triplets that are similar. The intuition is that to obtain representation of an unseen target triplet $t'=(person,ride,camel)$, we wish to use only similar triplets such as $t=(person,ride,horse)$ but not triplets such as $t=(person,ride,skateboard)$.
 For this, we propose to decompose the similarity between triplets $t$ and $t'$ by looking at the similarities between their subjects, predicates and objects measured by the dot-product of their representations in the corresponding individual embedding spaces. The motivation is that the subject, object and predicate spaces do not suffer as much from the limited training data compared to the visual phrase space. In detail, we define a weighting function $G$ as :  
\begin{align}
\label{eq:similarity}
G(t,t') = \sum_{b \in \{s,p,o\}} \alpha_b {\bm{w}^b_t}^T  \bm{w}^b_{t'},
\end{align}
where ${\bm{w}^b_t}^T  \bm{w}^b_{t'}$ measures similarity between embedded representations $\bm{w}^b_{.}$ and scalars $\alpha_b$ are hyperparameters that reweight the relative contribution of subject, object and predicate similarities. As we constrain $\sum_b \alpha_b = 1$ the output of $G(t,t') \in [0,1]$. For a target triplet $t'$, we define as $\mathcal{N}_{t'}$ the set of $k$ most similar source triplets according to $G$. 

\vspace{-2.5ex}
\paragraph{Learning $\Gamma$.}
We fit parameters of $\Gamma$ by learning analogy transformations between triplets in the training data. In particular, we generate training data pairs of source $t$ and target $t'$ triplets. 
Given the generated data, we optimize log-likelihood similar to Eq.~\eqref{eq:loss} but using visual features of the real target triplet and language features of the source triplet transformed with the analogy transformation $\Gamma$.  The optimization is performed w.r.t. to both the parameters of $\Gamma$ and parameters of the embedding functions. Details are given in the Section~\ref{part:supmat_optim_2} of the Appendix. 

\vspace{-2.5ex}
\paragraph{Aggregating embeddings.}
At test time, we compute the visual phrase embedding of an unseen triplet $u$ by aggregating embeddings of similar seen triplets $t \in \mathcal{N}_u$ transformed using the analogy transformation: 
\begin{equation}
\label{eq:aggregation}
\bm{\bar{w}}^{vp}_{u} = \sum_{t \in \mathcal{N}_u} G(t,u)~(\bm{w}^{vp}_{t} + \Gamma(t,u)),
\end{equation}
where $\bm{w}^{vp}_{t}$ is the visual phrase embedding of source triplet $t$ obtained with Eq.~\eqref{eq:proj2}, $\Gamma(t,u)$ is the analogy transformation between source triplet $t$ and unseen triplet $u$ computed by Eq.~\eqref{eq:gamma_reg} and $G(t,u)$ is a scalar weight given by Eq.~\eqref{eq:similarity} that re-weights the contribution of the different source triplets. This process is illustrated in Figure~\ref{fig:model}(b). 

%% file: experiments_fig_hico.tex
\begin{figure*}[t]
\vspace{-0.7cm}
\centering
	\begin{minipage}[t]{0.18\textwidth}
    \centering
    	\textit{Query (Q) / Source (S)}\\
    	\vspace{0.2ex}
	\end{minipage}	
	\hspace{0.01\textwidth}
	\begin{minipage}[t]{0.58\textwidth}
    	\centering
    \textit{	Top true positives}\\
    	\vspace{0.2ex}
	\end{minipage}
	\hspace{0.005\textwidth}
	\begin{minipage}[t]{0.18\textwidth}
    \centering
    \textit{	Top false positive}\\
    	\vspace{0.2ex}
	\end{minipage}

    \begin{minipage}[c]{0.20\textwidth}
    \vspace{-10ex}
    		\small{
	(Q) \textbf{{\color{blue}person} {\textcolor{Green}{pet}} {\color{red}cat}}\\
	\vspace{-18pt}
	\par\noindent\rule{\textwidth}{0.4pt}
	(S) {\color{blue}person} {\textcolor{Green}{pet}} {\color{red}dog} \\
	(S) {\color{blue}person} {\textcolor{Green}{pet}} {\color{red}giraffe} \\
	(S) {\color{blue}person} {\textcolor{Green}{pet}} {\color{red}cow} \\
	(S) {\color{blue}person} {\textcolor{Green}{pet}} {\color{red}elephant} \\
	(S) {\color{blue}person} {\textcolor{Green}{scratch}} {\color{red}cat} \\
		}
    \end{minipage}
    \hspace{0.005\textwidth} 
    \begin{minipage}[t]{0.18\textwidth}
    	\centering
       	\includegraphics[trim={1.3cm 1cm 1cm 1cm},clip,width=0.95\linewidth,cfbox={green 2pt 2pt}]{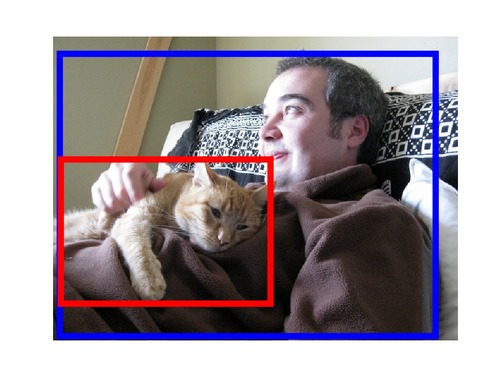}\\
       	\vspace{1.5ex}
    \end{minipage}
    \hspace{0.005\textwidth}
    \begin{minipage}[t]{0.18\textwidth}
    	\centering
       	\includegraphics[trim={0.2cm 1.2cm 0cm 1.1cm},clip,width=0.95\linewidth,cfbox={green 2pt 2pt}]{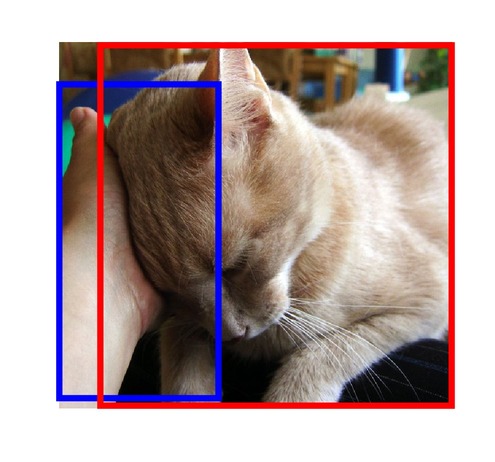}\\
       	\vspace{1.5ex}
    \end{minipage}
    \hspace{0.005\textwidth}
    \begin{minipage}[t]{0.18\textwidth}
       \centering
       \includegraphics[trim={0.2cm 2cm 0cm 3.5cm},clip,width=0.95\linewidth,cfbox={green 2pt 2pt}]{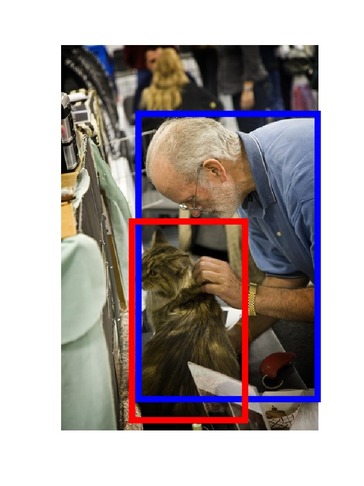}\\
       \vspace{1.5ex}
    \end{minipage}
    \hspace{0.005\textwidth}
    \begin{minipage}[t]{0.18\textwidth}
    	\centering
       	\includegraphics[trim={1.2cm 1cm 2.3cm 1cm},clip,width=0.95\linewidth,cfbox={red 2pt 2pt}]{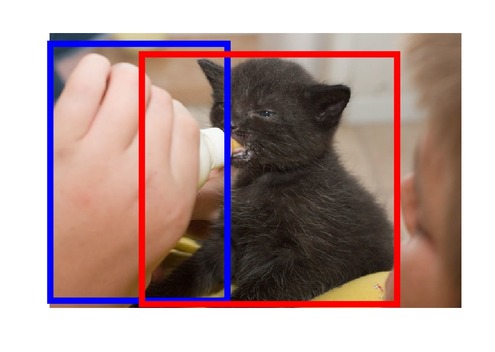}\\
      	\vspace{1.5ex}
    \end{minipage}

    \setlength\abovecaptionskip{1pt}
    \caption{\small Top retrieved positive (green) and negative (red) detections with our model (s+o+vp+transfer) on unseen triplets excluded from HICO-DET. 
For a target triplet (Q) (e.g. ``person pet cat"), our model automatically learns to select meaningful source triplets (S) involving visually similar objects or predicates (``person pet dog", ``person scratch cat") and transforms their visual phrase embeddings by analogy transformation $\Gamma$. The top false positive corresponds to a visually related action (``feed"). Additional examples are in Section~\ref{part:supmat_hico} of the Appendix.}
    \label{fig:qualitative_hico}
    \vspace*{-5mm}
\end{figure*}

%% file: experiments.tex

In this section we evaluate the performance of our model for visual relation retrieval on three challenging datasets : HICO-DET~\cite{Chao18}, UnRel~\cite{Peyre17} and COCO-a~\cite{Ronchi2015}.
Specifically, we numerically assess the two components of our model : (i) learning the visual phrase embedding together with the unigram embeddings and (ii) transferring embeddings to unseen triplets by analogy transformations.

\subsection{Datasets and evaluation set-ups}
\label{part:datasets}

\paragraph{HICO-DET.}
The HICO-DET~\cite{Chao15, Chao18} dataset contains images of human-object interactions with box-level annotations. The interactions are varied : the vocabulary of objects matches the 80 COCO~\cite{Lin2014a} categories and there are 117 different predicates. The number of all possible triplets is $1 \times 117 \times 80$ but the dataset contains positive examples for only 600 triplets. All triplets are seen at least once in training. The authors separate a set of 138 rare triplets, which are the triplets that appear fewer than 10 times at training. To conduct further analysis of our model, we also select a set of 25 triplets that we treat as unseen, exclude them completely from the training data in certain experiments, and try to retrieve them at test time using our model. These triplets are randomly selected among the set of non-rare triplets in order to have enough test instances on which to reliably evaluate. 

\vspace{-.3cm}
\paragraph{UnRel.}
UnRel~\cite{Peyre17} is an evaluation dataset containing visual relations for 76 unusual triplet queries. In contrast to HICO-DET and COCO-a, the interactions do not necessarily involve a human, and the predicate is not necessarily an action (it can be a spatial relation, or comparative). The vocabulary of objects and predicates matches those of Visual Relation Detection Dataset~\cite{Lu16}. UnRel is only an evaluation dataset, so similar to~\cite{Peyre17} we use the training set of Visual Relationship Dataset as training data. 

\input{experiments_fig_unrel}

\vspace{-.3cm}
\paragraph{COCO-a.}
The COCO-a dataset~\cite{Ronchi2015} is based on a subset of COCO dataset~\cite{Lin2014a} augmented with annotations of human-object interactions. Similar to HICO-DET, the vocabulary of objects matches the 80 COCO categories. In addition, COCO-a defines 140 predicates resulting in a total of 1681 different triplets. %
The released version of COCO-a contains 4413 images with no pre-defined train/test splits. Given this relatively small number of images, we use COCO-a as an evaluation dataset for models trained on HICO-DET. This results in an extremely challenging set-up with 1474 unseen triplets among which 1048 involve an out-of-vocabulary predicate that has not been seen at training in HICO-DET.

\vspace{-.35cm}
\paragraph{Evaluation measure.}
On all datasets, we evaluate our model in a retrieval setup. For each triplet query in the vocabulary, we rank the candidate test pairs of object bounding boxes using our model and compute the performance in terms of Average Precision. Overall, we report mean Average Precision (mAP) over the set of triplet queries computed with the evaluation code released by~\cite{Chao18} on HICO-DET and~\cite{Peyre17} on UnRel. On COCO-a, we use our own implementation as no evaluation code is released. 

\subsection{Implementation details} 
\label{part:implementation_details}

\paragraph{Candidate pairs.}
We use pre-extracted candidate pairs of objects from an object detector trained for the vocabulary of objects specific to the dataset. On HICO-DET, we train the object detector on the COCO training data using Detectron~\cite{Detectron2018}. To be comparable to~\cite{Gkioxari18}, we use a Faster-R-CNN \cite{ren15} with ResNet-50 Feature Pyramid Network \cite{Lin2017FeaturePN}. We post-process the candidate detections by removing candidates whose confidence scores are below 0.05 and apply an additional per-class score thresholding to maintain a fixed precision of 0.3 for each object category. At test time, we use non-maximum suppression of 0.3. 
For COCO-a, we re-train the object detector excluding images from COCO that intersect with COCO-a. On UnRel, we use the same candidate pairs as~\cite{Peyre17} to have directly comparable results.

\vspace{-.3cm}
\paragraph{Visual representation.} 
Following~\cite{Peyre17}, we first encode a candidate pair of boxes $(\bm{o_s},\bm{o_o})$ by the appearance of the subject $\bm{a}(\bm{o_s})$, the appearance of the object $\bm{a}(\bm{o_o})$, and their mutual spatial configuration $\bm{r}(\bm{o_s},\bm{o_o})$. The appearance features of the subject and object boxes are extracted from the last fully-connected layer of the object detector. The spatial configuration $\bm{r}(\bm{o_s},\bm{o_o})$ is a 8-dimensional feature that concatenates the subject and object box coordinates renormalized with respect to the union box. The visual representation of a candidate pair is a 1000-dimensional vector, aggregating the spatial and appearance features of the objects (more details in Section~\ref{part:supmat_optim_1} of the Appendix). For the subject (resp. object) embeddings, we only consider the appearance of the subject (resp. object) without the spatial configuration.

\vspace{-.3cm}
\paragraph{Language representation.} 
For a triplet $t=(s,p,o)$, we compute the word embeddings $e_s$ (resp. $e_p$, $e_o$) for subject (resp. predicate, object) with a Word2vec~\cite{mikolov2013distributed} model trained on GoogleNews. The representation of a triplet is taken as the concatenation of the word embeddings $\bm{\mathbf{q}}_t = [\bm{e}_s; \bm{e}_p; \bm{e}_o] \in \mathbb{R}^{900}$. 

\vspace{-.3cm}
\paragraph{Embedding functions.} The embedding projection functions are composed of two fully connected layers, with a ReLU non-linearity. For the visual projection functions, we use Dropout. The dimensionality of the joint visual-language spaces is set to $d=1024$ for HICO-DET and COCO-a. We use $d=256$ for UnRel as the training set is much smaller. 

\vspace{-.3cm}
\paragraph{Training details.} 
We train our model with Adam optimizer~\cite{Adam} using a learning rate 0.001. We first learn the parameters of the projection functions by optimizing $\mathcal{L}_{joint}$, then activate the analogy loss $\mathcal{L}_{\Gamma}$ to learn the parameters of transfer and finetune the visual phrase embeddings. The hyperparameters $\alpha_s$, $\alpha_o$, $\alpha_p$ and $k$ are optimized by grid-search on the validation set. More details on optimization and batch sampling are provided in Section~\ref{part:supmat_optim_1} of the Appendix. 

\begin{table}[b]
\vspace{-3ex}
\centering
\small{
\begin{tabular}{@{}rrrr@{}}\toprule
											& full 	& rare 	& non-rare \\\midrule
Chao \cite{Chao18} 							&  7.8 	& 5.4 	& 8.5	\\
Gupta \cite{Gupta15} 						&  9.1 	& 7.0	& 9.7  \\
Gkioxari \cite{Gkioxari18} 					&  9.9	& 7.2	& 10.8 	\\
GPNN \cite{Qi18}								& 13.1 & 9.3 	& 14.2 \\
iCAN \cite{iCAN}								& 14.8 & 10.5 & 16.1 \\
\text{s+o+p} 								& 18.7 & 13.8	& 20.1 \\
\text{s+o+vp}  								& 17.7 & 11.6	& 19.5 \\
\text{s+o+p+vp}								& \textbf{19.4} 	& \textbf{14.6}	& \textbf{20.9} 	\\
\bottomrule
\end{tabular}
}
\vspace{-1ex}
\caption{\small Retrieval results on HICO-DET dataset (mAP).}
\label{tab:results_hico_map}
\end{table}

\subsection{Evaluating visual phrases on seen triplets}

\begin{table}[t!]
\setlength\tabcolsep{2.9pt} 
\centering
\small{
\ra{1}
\begin{tabular}{@{}rccccc@{}}\toprule
\vspace{0.5ex}
& \multicolumn{1}{c}{Base} & \multicolumn{4}{c}{With aggregation $G$}\\
& - & \hbox{\thickmuskip=1mu$\Gamma=\O$} & \hbox{\thickmuskip=1mu$\Gamma=0$}  & \hbox{\thickmuskip=1mu$\Gamma=linear$} & \hbox{\thickmuskip=1mu$\Gamma=deep$}
\\\midrule
\rule{0pt}{1ex} 				
\text{s+o+p}						& 23.2 	&	-	& -			& -			& - \\
\text{s+o+vp+transfer}			& 24.1	& 9.6	& 24.8 		&	27.6 	& \textbf{28.6}\\
\text{s+o+p+vp+transfer}			& 23.6	& 12.5	& 24.5 		&	25.4  	& \textbf{25.7} \\
\text{supervised}				& 33.7	&	-	& - 			&	-  		& - \\
\bottomrule
\end{tabular}
}
\vspace{-2mm}
\caption{\small mAP on the 25 zero-shot test triplets of HICO-DET with variants of our model trained on the $trainval$ set excluding the positives for the zero-shot triplets. The first column shows the results without analogy transfer (Section~\ref{part:model_1}) while the other columns display results with transfer using different forms of analogy transformation $\Gamma$ (Section~\ref{part:model_2}). Last line (supervised) is the performance of (s+o+p+vp) trained will all training instances.}
\label{tab:results_hico_zeroshot}
\vspace{-4mm}
\end{table}

\input{experiments_fig_cocoa}

We first validate the capacity of our model to detect triplets seen at training and compare with recent state-of-the-art methods. In Table~\ref{tab:results_hico_map}, we report mAP results on HICO-DET in the Default setting defined by~\cite{Chao18} on the different subsets of triplets (full), (rare), (non rare) as described in~\ref{part:datasets}. First, we compute three variants of our model : (i) the compositional part using all unigram terms (s+o+p), which can be viewed as a strong fully compositional baseline, (ii) the visual phrase part combined with object scores (s+o+vp), and (iii) our full model (s+o+p+vp) that corresponds to the addition of the visual phrase representation on top of the compositional baseline (section~\ref{part:model_1}). The results show that our visual phrase embedding is beneficial, leading to a consistent improvement over the strong compositional baseline on all sets of triplets, improving the current state-of-the art~\cite{iCAN} by more than 30\% in terms of relative gain. We provide ablation studies in Section~\ref{part:supmat_ablation} of the Appendix as well as experiments incorporating bigrams modules (sr+ro) leading to improved results.

\subsection{Transfer by analogy on unseen triplets}
Next, we evaluate the benefits of transfer by analogy focusing on the challenging set-up of triplets never seen at training time. While the HICO-DET dataset contains both seen (evaluated in previous section) and manually constructred unseen triplets (evaluated here), in this section we consider additional two datasets that contain only unseen triplets. In particular, we use UnRel to  evaluate retrieval of unusual (and unseen) triplets and COCO-a to evaluate retrieval of unseen triplets with out-of-vocabulary predicates.
 
\vspace{-.35cm}
\paragraph{Evaluating unseen triplets on HICO-DET.} 
First, we evaluate our model of transfer by analogy on the 25 zero-shot triplets of HICO-DET. In Table~\ref{tab:results_hico_zeroshot}, we show results for different types of analogy transformations applied to the visual phrase embeddings to be compared with the base model not using analogy (first column). 
First, \hbox{\thickmuskip=1mu$\Gamma=\O$} corresponds to aggregation of visual phrase embeddings of source triplets without analogy transformation. Then, we report three variants of an analogy transformation, where visual phrase embeddings are trained with analogy loss and the embedding of source triplet is either (i) aggregated without transformation (\hbox{\thickmuskip=1mu$\Gamma=0$}), or transformed with (ii) a linear transformation (\hbox{\thickmuskip=1mu$\Gamma=linear$}) or (iii) a 2-layer perceptron (\hbox{\thickmuskip=1mu$\Gamma=deep$}).
The results indicate that forming visual phrase embeddings of unseen test triplets by analogy transformations of similar seen triplets, as described in~\ref{part:model_2}, is beneficial, with the best model (s+o+vp+transfer using \hbox{\thickmuskip=1mu$\Gamma=deep$}) providing a significant improvement over the compositional baseline (from mAP of 23.2 to 28.6), thus partly filling the gap to the fully supervised setting (mAP of 33.7). It is also interesting to note that, when aggregating visual phrase embeddings of different source triplets as described in Eq.~\eqref{eq:aggregation},  transforming the visual phrase embedding via analogy prior to the aggregation is necessary, as indicated by the significant drop of performance when \hbox{\thickmuskip=1mu$\Gamma=\O$}. 
In Figure~\ref{fig:qualitative_hico} we show qualitative results for retrieval of unseen triplets with the (s+o+vp+transfer) model. For a query triplet (Q) such as ``person pet cat" we show the top 3 retrieved candidate pairs (green), and the top 1 false positive (red). Also, for each target triplet, we show the source triplets (S) used in the transfer by analogy (Eq.~\eqref{eq:aggregation}). We note that the source triplets appear relevant to the query. 

\begin{table}[b]
\vspace{-2.5ex}
\centering
\small{
\ra{1}
\begin{tabular}{@{}rrrrrr@{}}\toprule
& \multicolumn{1}{c}{With GT} & \multicolumn{3}{c}{With candidates} \\
& - & union & subj & subj/obj
\\\midrule
DenseCap \cite{Johnson2015} 		& - 		& 6.2 	& 6.8 	& - \\
Lu \cite{Lu16}					& 50.6 	& 12.0 	& 10.0 	& 7.2  \\
Peyre \cite{Peyre17} full 		& 62.6 	& 14.1 	& 12.1 	& 9.9 \\
\text{p}							& 62.2	& 16.8	& 15.2	& 12.6	 	 \\
\text{vp}						& 53.4	& 13.2	& 11.7	& 9.4 			\\
\text{p+vp}						& 61.7	& 16.4	& 14.9	& 12.6 \\
\text{vp+transfer}				& 53.7	& 13.7	& 12.0 	& 9.7				\\
\text{p+vp+transfer}				& \textbf{63.9}	& \textbf{17.5}	&\textbf{15.9}	& \textbf{13.4}			\\
\bottomrule
\end{tabular}
\setlength\abovecaptionskip{5pt}
\caption{\small Retrieval on UnRel (mAP) with IoU=0.3.} 
\label{tab:results_unrel}
}
\vspace{-6mm}
\end{table}

\vspace{-.3cm}
\paragraph{Evaluating unseen (unusual) triplets on UnRel.}
Table~\ref{tab:results_unrel} shows numerical results for retrieval on the UnRel dataset. Similar to~\cite{Peyre17}, we also do not use subject and object scores as we found them uninformative on this dataset containing hard to detect objects. For transfer by analogy we use \hbox{\thickmuskip=1mu$\Gamma=deep$}.
First, we observe that our (p+vp+transfer) method improves over all other methods, significantly improving the current state-of-the-art~\cite{Peyre17} on this data, as well as outperforming the image captioning model of~\cite{Johnson2015} trained on a larger corpus. Note that we use the same detections and features as~\cite{Peyre17}, making our results directly comparable. 
Second, the results confirm the benefits of transfer by analogy (p+vp+transfer) over the fully compositional baseline (p) with a consistent improvement in all evaluation metrics. Interestingly, contrary to HICO-DET, using visual phrase embeddings without transfer (p+vp) does not bring significant improvements over (p). This is possibly due to the large mismatch between training and test data as the UnRel dataset used for testing contains unusual relations, as shown in the qualitative examples in Figure~\ref{fig:qualitative_unrel}. This underlines the importance of the transfer by analogy model. 

\begin{table}[t!]
\centering
\small{
\begin{tabular}{@{}rcccc@{}}\toprule
						& all		& out of vocabulary  \\\midrule
\text{s+o+p}				&	4.3 	& 	4.2	 \\
\text{s+o+vp}			& 	6.0		& 	6.2 \\
\text{s+o+p+vp}			&	5.1 	& 	5.1		 \\
\text{s+o+vp+transfer}	& 	\textbf{6.9}		&  	\textbf{7.3}		\\
\text{s+o+p+vp+transfer}	&	5.2 	& 	5.1 		 \\
\bottomrule
\end{tabular}
}
\vspace*{-2mm}
\caption{\small Retrieval on unseen triplets of COCO-a (mAP). We show the performance on all unseen triplets (first column) and on unseen triplets involving out-of-vocabulary predicates (second column).}
\label{tab:results_cocoa}
\vspace{-4mm}
\end{table}

\vspace{-.3cm}
\paragraph{Evaluating unseen (out-of-vocabulary) triplets on COCO-a.}
Finally, we evaluate our model trained on HICO-DET dataset for retrieval on the unseen triplets of COCO-a dataset. This is an extremely challenging setup as the unseen triplets of COCO-a involve predicates that are out of the vocabulary of the training data. The results shown in Table~\ref{tab:results_cocoa} demonstrate the benefits of the visual phrase representation as previously observed on HICO-DET and UnRel datasets. Furthermore, the results also demonstrate the benefits of analogy transfer : compared to the fully compositional baseline (s+o+p) our best analogy model (s+o+vp+transfer) obtains a relative improvement of 60\% on all, and more than 70\% on the out of vocabulary triplets. Qualitative results are shown in Figure~\ref{fig:qualitative_cocoa}.

%% file: experiments_fig_unrel.tex
\begin{figure*}[t]
\vspace{-0.3cm}
\centering
	\begin{minipage}[t]{0.18\textwidth}
    \centering
    	\textit{Query (Q) / Source (S)}\\
    	\vspace{0.2ex}
	\end{minipage}	
	\hspace{0.01\textwidth}
	\begin{minipage}[t]{0.58\textwidth}
    	\centering
    \textit{	Top true positives}\\
    	\vspace{0.2ex}
	\end{minipage}
	\hspace{0.005\textwidth}
	\begin{minipage}[t]{0.18\textwidth}
    \centering
    \textit{	Top false positive}\\
    	\vspace{0.2ex}
	\end{minipage}

    \begin{minipage}[c]{0.20\textwidth}
    \vspace{-12ex}	  
    	\small{
	(Q) \textbf{{\color{blue}dog} {\textcolor{Green}{wear}} {\color{red}shoes}} \\ 
	\vspace{-18pt}
	\par\noindent\rule{\textwidth}{0.4pt}
	(S) {\color{blue}person} {\textcolor{Green}{wear}} {\color{red}shoes} \\
	(S) {\color{blue}person} {\textcolor{Green}{wear}} {\color{red}shoe} \\
	(S) {\color{blue}person} {\textcolor{Green}{wear}} {\color{red}skis}\\
	(S) {\color{blue}person} {\textcolor{Green}{wear}} {\color{red}pants} \\
	(S) {\color{blue}person} {\textcolor{Green}{wear}} {\color{red}jeans}\\
	}
    \end{minipage}  
    \hspace{0.005\textwidth} 
    \begin{minipage}[t]{0.18\textwidth}
    	\centering
       	\includegraphics[trim={2.4cm 0.7cm 2cm 0.5cm},clip,width=0.95\linewidth,cfbox={green 2pt 2pt}]{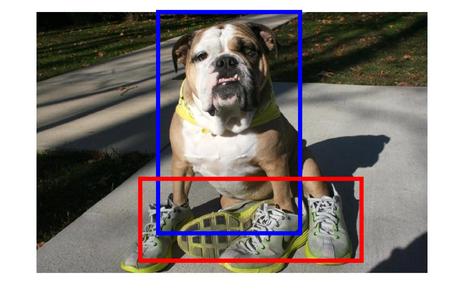}\\
       	\vspace{1.5ex}
    \end{minipage}
    \hspace{0.005\textwidth}
    \begin{minipage}[t]{0.18\textwidth}
    	\centering
       	\includegraphics[trim={1.9cm 1cm 1.4cm 1.1cm},clip,width=0.95\linewidth,cfbox={green 2pt 2pt}]{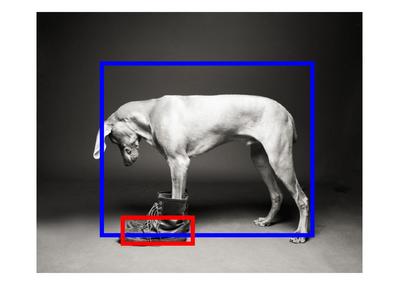}\\
       	\vspace{1.5ex}
    \end{minipage}
    \hspace{0.005\textwidth}
    \begin{minipage}[t]{0.18\textwidth}
       \centering
       \includegraphics[trim={0.5cm 1.1cm 0.5cm 1cm},clip,width=0.95\linewidth,cfbox={green 2pt 2pt}]{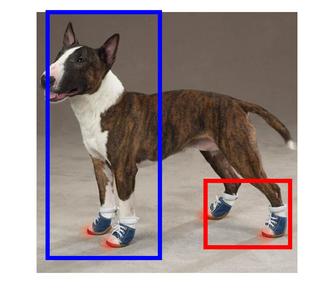}\\
       \vspace{1.5ex}
    \end{minipage}
    \hspace{0.005\textwidth}
    \begin{minipage}[t]{0.18\textwidth}
    	\centering
       	\includegraphics[trim={0cm 0.9cm 0cm 0.75cm},clip,width=0.95\linewidth,cfbox={red 2pt 2pt}]{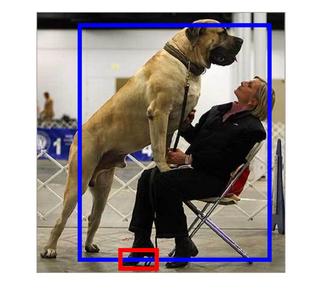}\\
      	\vspace{1.5ex}
    \end{minipage}   
    
    \setlength\abovecaptionskip{1pt}
    \caption{\small Top retrieved positive (green) and negative (red) detections with our model (p+vp+transfer) on UnRel triplets. The embedding of the unseen query triplet (Q) is formed from the embedding of seen source triplets (S) via analogy transformation. While transfer with analogy on HICO-DET is often done through change of object, here, for retrieving the unseen triplet ``dog wear shoes", our model samples source triplets involving a different subject, ``person", in interaction with similar objects (e.g. ``person wear shoes", ''person wear skis"). Additional examples are in Section~\ref{part:supmat_unrel} of the Appendix.}
    \label{fig:qualitative_unrel}
     \vspace*{-5mm}
\end{figure*}

%% file: experiments_fig_cocoa.tex
\begin{figure*}[t]
\vspace{-0.3cm}
\centering
	\begin{minipage}[t]{0.18\textwidth}
    \centering
    	\textit{Query (Q) / Source (S)}\\
    	\vspace{0.2ex}
	\end{minipage}	
	\hspace{0.01\textwidth}
	\begin{minipage}[t]{0.58\textwidth}
    	\centering
    \textit{	Top true positives}\\
    	\vspace{0.2ex}
	\end{minipage}
	\hspace{0.005\textwidth}
	\begin{minipage}[t]{0.18\textwidth}
    \centering
    \textit{	Top false positive}\\
    	\vspace{0.2ex}
	\end{minipage}

    \begin{minipage}[c]{0.20\textwidth}
    \vspace{-8ex}	  
    	\small{
	(Q) \textbf{{\color{blue}person} {\textcolor{Green}{taste}} {\color{red}cup}} \\ 
	\vspace{-18pt}
	\par\noindent\rule{\textwidth}{0.4pt}
	(S) {\color{blue}person} {\textcolor{Green}{fill}} {\color{red}cup} \\
	(S) {\color{blue}person} {\textcolor{Green}{smell}} {\color{red}cup} \\
	(S) {\color{blue}person} {\textcolor{Green}{cook}} {\color{red}hot dog}\\
	(S) {\color{blue}person} {\textcolor{Green}{make}} {\color{red}vase} \\
	(S) {\color{blue}person} {\textcolor{Green}{cut}} {\color{red}apple}\\
	}
    \vspace{2ex}
    \end{minipage} 
    \hspace{0.005\textwidth} 
    \begin{minipage}[t]{0.18\textwidth}
    	\centering
       	\includegraphics[trim={1.2cm 0.5cm 1cm 0.35cm},clip,width=0.95\linewidth,cfbox={green 2pt 2pt}]{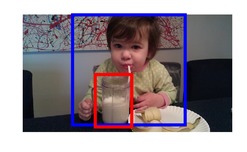}\\
       	\vspace{1.5ex}
    \end{minipage}
    \hspace{0.005\textwidth}
    \begin{minipage}[t]{0.18\textwidth}
    	\centering
       	\includegraphics[trim={1.75cm 1.5cm 0.4cm 0.4cm},clip,width=0.95\linewidth,cfbox={green 2pt 2pt}]{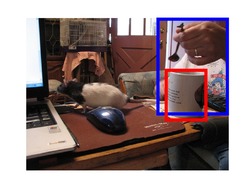}\\
       	\vspace{1.5ex}
    \end{minipage}
    \hspace{0.005\textwidth}
    \begin{minipage}[t]{0.18\textwidth}
       \centering
       \includegraphics[trim={-0.5cm 0.7cm -0.6cm 1.3cm},clip,width=0.95\linewidth,cfbox={green 2pt 2pt}]{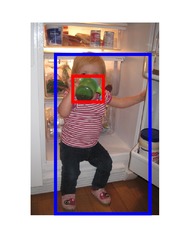}\\
       \vspace{1.5ex}
    \end{minipage}
    \hspace{0.005\textwidth}
    \begin{minipage}[t]{0.18\textwidth}
    	\centering
       	\includegraphics[trim={0.7cm 0.75cm 0.5cm 0.5cm},clip,width=0.95\linewidth,cfbox={red 2pt 2pt}]{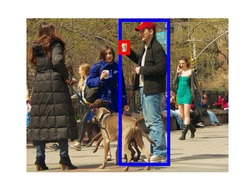}\\
      	\vspace{1.5ex}
    \end{minipage}    
    \vspace*{-4mm}
    \setlength\abovecaptionskip{1pt}
    \caption{\small Top retrieved positives (green) and negatives (red) detections with our model (s+o+vp+transfer) of COCO-a triplets. The embedding of the query triplet (Q) to retrieve is formed with the embedding of source triplets (S) by analogy. For retrieving out-of-vocabulary triplets such as ``person taste cup", our model of transfer by analogy automatically samples relevant source triplets involving similar predicates and objects (e.g. ``person smell cup", ``person make vase"). Additional examples are in Section~\ref{part:supmat_cocoa} of the Appendix.}
    \label{fig:qualitative_cocoa}
    \vspace*{-3mm}
\end{figure*}

%% file: conclusion.tex
We have developed a new approach for visual relation detection that combines compositional and visual phrase representations.
Furthermore, we have proposed a model for transfer by analogy able to compute visual phrase embeddings of never seen before relations. We have demonstrated benefits of our approach on three challenging datasets involving unseen triplets.





%% file: appendix.tex
\renewcommand{\thetable}{\Alph{table}}
\renewcommand{\thefigure}{\Alph{figure}}
\setcounter{figure}{0}   
\setcounter{table}{0}

\section*{Overview}
In this Appendix, we provide (i) technical details of our model described in Section~\ref{part:model} of the main paper (Section~\ref{part:supmat_optim}), (ii) additional ablation studies to better understand the benefits of the different components of our model (Section~\ref{part:supmat_ablation}), (iii) additional qualitative results of our model for transfer by analogy on HICO-DET dataset (Section~\ref{part:supmat_hico}), UnRel dataset (Section~\ref{part:supmat_unrel}) and COCO-a dataset (Section~\ref{part:supmat_cocoa}), and (iv) a qualitative analysis of joint embedding spaces by the t-sne~\cite{tsne} visualization (Section~\ref{part:supmat_tsne}).

\section{Additional details of our model}
\label{part:supmat_optim}

In this part, we provide additional details of our model that we could not include in the main paper due to space constraints. 
First,  in Section~\ref{part:supmat_optim_2}, we describe how we learn the analogy transformation including details on (i) sampling source triplets, (ii) training loss and (iii) optimization. Second, in Section~\ref{part:supmat_optim_1}, we detail our visual representation and explain how we form mini-batches during training.

\subsection{Learning analogy transformations}
\label{part:supmat_optim_2}

\paragraph{Sampling source triplets.} Please recall (Section~\ref{part:model_2} of the main paper) that we fit parameters of $\Gamma$ by learning analogy transformations between triplets available in the training data. To do this, we generate pairs of source $t$ and
target $t'$ triplets as follows. For a target triplet $t'$ in the training data, the source triplets for transfer by analogy are sampled in two steps : (i) for a given target triplet $t'$, we first compute the similarity $G(t,t')$ given by Eq.~\eqref{eq:similarity} using all triplets $t$ in the training data that occur at least 10 times (i.e. the non-rare triplets according to the definition of~\cite{Chao18}), (ii) we sort this set of candidate source triplets, and retain the top k most similar triplets according to $G$. The outcome is a set of source triplets $\mathcal{N}_{t'}$, similar to the target triplet $t'$ and hence suitable for learning the analogy transformation. 
Please note that we do not constrain the source triplets to share words with the target triplet, so all words may differ between source and target triplets. Also note that the procedure described above is similar at training and test time. 
In practice, we take $k=5$, $\alpha_r=0.8$, $\alpha_s=\alpha_o=0.1$ for all datasets. These hyperparameters are optimized by grid-search on the validation set of HICO-DET.

\begin{figure}[t]
\includegraphics[trim={3.5cm 3.8cm 3.5cm 3cm},clip,width=\linewidth]{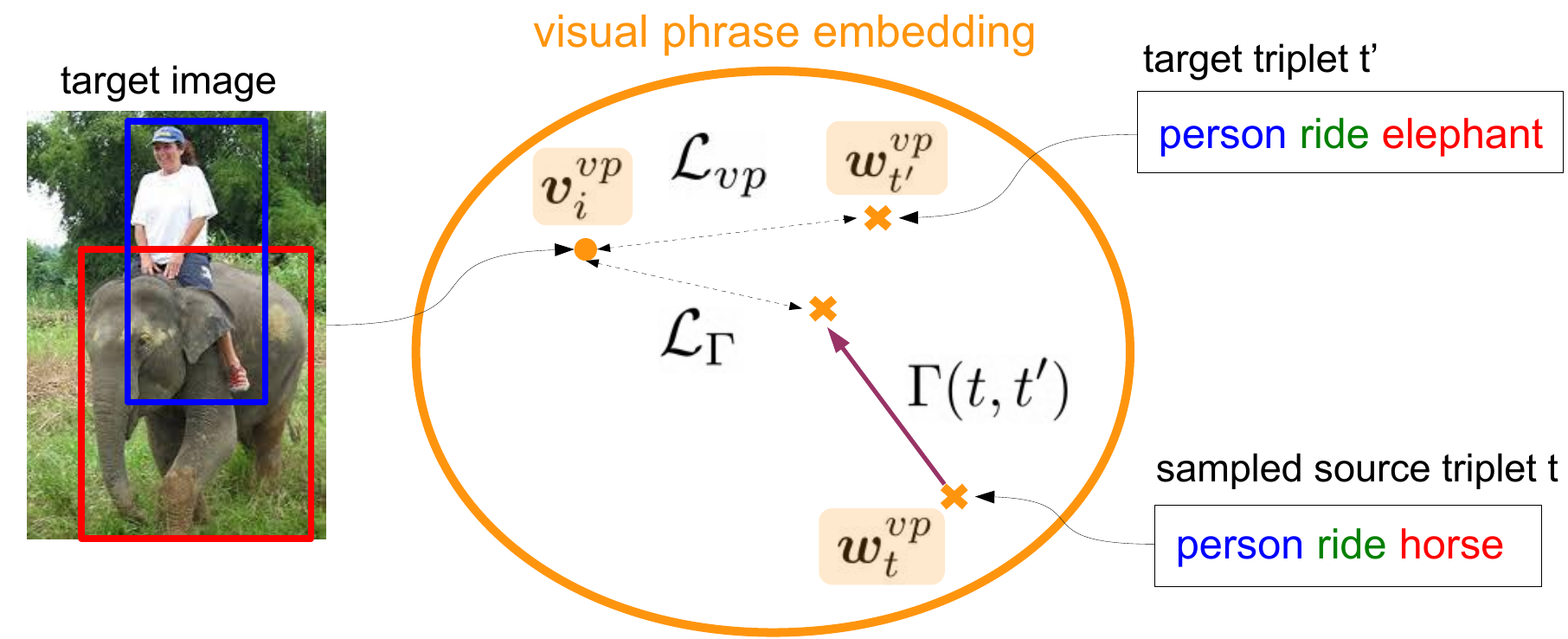}
\vspace{-2ex}
\caption{\small Illustration of training the analogy transformation $\Gamma$. For each target triplet $t'$ (e.g. ``person ride elephant"), we randomly sample a source triplet $t$ (e.g. ``person ride horse"). The first part of the analogy loss $\mathcal{L}_{\Gamma}$ in Eq~\eqref{eq:loss_gamma} encourages that the transformed visual phrase embedding $w_t^{vp}+\Gamma(t,t')$ is close to the corresponding visual representation $v_i^{vp}$ of target triplet $t$.}
\vspace{-2ex}
\label{fig:analogy_train}
\end{figure}

\vspace{-0.3cm}
\paragraph{Learning $\Gamma$.}
For each target triplet $t'$ in the training batch, we randomly sample a relevant source triplet $t \in \mathcal{N}_{t'}$ as described above. We call $\mathcal{Q}$ the set of pairs of related triplets $(t,t')$ formed like this. The parameters of $\Gamma$ are learnt by maximizing the log-likelihood : 
\vspace{-0.2cm}
\begin{align}
\label{eq:loss_gamma}
\mathcal{L}_{\Gamma} =& \sum_{i=1}^N \sum_{(t,t') \in \mathcal{Q}} \mathbb{1}_{y_{t'}^i=1} \log \left( \frac{1}{1+e^{-{(\bm{w}^{vp}_{t} + \Gamma(t,t'))}^{T} \bm{v}_i^{vp}}} \right) \notag \\
 +& \sum_{i=1}^N \sum_{(t,t') \in \mathcal{Q}} \mathbb{1}_{y_{t'}^i=0} \log \left( \frac{1}{1+e^{{(\bm{w}^{vp}_{t} + \Gamma(t,t'))}^T \bm{v}_i^{vp}}} \right),
\end{align}\\[-.3cm]
where $\bm{v}_i^{vp}$ are the visual features projected to the visual phrase space and $(\bm{w}^{vp}_{t} + \Gamma(t,t'))$ is the transformed visual phrase embedding of the source triplet $t$ to target triplet $t'$ following Eq.~\eqref{eq:gamma} in the main paper. Note that this loss is similar to the loss used for learning embeddings of visual relations, given by Eq.~\eqref{eq:loss} in the main paper. The first attraction term pulls closer visual representation $\bm{v}_i^{vp}$ to its corresponding language representation $\bm{w}^{vp}_{t} + \Gamma(t,t')$ obtained via analogy transformation, i.e. where the visual representation matches the embedding of the target triplet $t'$ obtained via analogy transformation. We illustrate this term in Figure~\ref{fig:analogy_train}. The second repulsive term pushes apart visual-language pairs that do not match, i.e. where the visual representation does not match the target triplet $t'$ obtained via the analogy transformation. 
The main idea behind Eq.~\eqref{eq:loss_gamma} is to use the analogy transformation $\Gamma$ to make the link between the language embedding of a source triplet $t$ and the visual embedding of a target triplet $t'$ in the joint $vp$ space. For example, let us consider source-target pairs of triplets $\mathcal{Q}=\{(t_1,t_1'),(t_2,t_2')\}$ in a mini-batch, where, $t_1=(person,ride,horse)$, $t_1'=(person,ride,elephant)$, $t_2=(person,pet,cat)$, $t_2'=(person,pet,sheep)$. The analogy loss in Eq.~\eqref{eq:loss_gamma} learns $\Gamma$ that transforms, in the joint $vp$ space, the language embedding of the source triplet $(person,ride,horse)$ such that it is close to the visual embedding of the target triplet $(person,ride,elephant)$ (first term in the loss) but far from the visual embedding of the other target triplet $(person,pet,sheep)$ (second term in the loss).

\vspace{-0.3cm}
\paragraph{Optimization details.}
First, we learn the parameters of embedding functions by optimizing $\mathcal{L}_{joint} = \mathcal{L}_s + \mathcal{L}_o + \mathcal{L}_p + \mathcal{L}_{vp}$ (Eq.~\eqref{eq:loss} in the main paper) for 10 epochs with Adam optimizer~\cite{Adam} using a learning rate 0.001. Then, we fix parameters of the embedding functions for $s$,~$o$ and $p$ and only finetune parameters of the visual phrase embedding function $vp$ while learning parameters of analogy transformation $\Gamma$. This is done by jointly optimizing $\mathcal{L}_{vp} + \lambda \mathcal{L}_{\Gamma}$ for 5 epochs with Adam optimizer~\cite{Adam} using a learning rate 0.001. In practice, we take $\lambda=1$. In this joint optimization, we found it helpful to restrict back-propagation of gradients coming from $\mathcal{L}_{\Gamma}$ only to the parameters of analogy transformation $\Gamma$ and parameters of the visual embedding functions $f^b_v$ (Eq.~\eqref{eq:proj}), i.e. we exclude back-propagation of gradients coming from $\mathcal{L}_{\Gamma}$ to parameters of language embedding functions $f^b_w$. These parameters are finetuned using gradients back-propagated from $\mathcal{L}_{vp}$.

\subsection{Implementation details}
\label{part:supmat_optim_1}

\paragraph{Visual representation.}
As described in Section~\ref{part:experiments} of the main paper, a candidate pair of bounding boxes $(\bm{o_s},\bm{o_o})$ is encoded by the appearance of the subject $\bm{a}(\bm{o_s})$, the appearance of the object $\bm{a}(\bm{o_o})$, and their mutual spatial configuration $\bm{r}(\bm{o_s},\bm{o_o})$. The spatial configuration $\bm{r}(\bm{o_s},\bm{o_o})$ is a 8-dimensional feature that concatenates the subject and object box coordinates renormalized with respect to the union box, i.e. we concatenate $[\frac{x_{min}-T}{A}, \frac{x_{max}-T}{A}, \frac{y_{min}-T}{A}, \frac{y_{max}-T}{A}]$ for subject and object boxes where $T$ and $A$ are the origin and the area of the union box, respectively. The visual representation of a candidate pair is then
\begin{align}
\bm{\mathbf{x}}_i = \begin{bmatrix}
						MLP_s(\bm{a}(\bm{o_s})) \\
						MLP_o(\bm{a}(\bm{o_o})) \\
						MLP_r(\bm{r}(\bm{o_s},\bm{o_o})) 
					\end{bmatrix},	
\end{align}
where $MLP_s$, $MLP_o$ contain one layer that projects the appearance features into a vector of dimension 300 and $MLP_r$ is a 2-layer perceptron projecting the spatial features into a vector of dimension 400, making the final dimension of $\bm{\mathbf{x}}_i$ equal to 1000. Note that both $p$ and $vp$ use the same visual input (including spatial features) while $s$ and $o$ modules only use the appearance features.

\vspace{-0.3cm}
\paragraph{Sampling batches.}
In practice, our model is trained with mini-batches containing 64 candidate object pairs. 25\% of the candidate pairs are positive, i.e. the candidate subject and object are interacting. The rest are negative, randomly sampled among candidate pairs involving the same subject and object category (but not interacting). For training, we use candidates from both ground truth and object detector outputs. At test time, we only use candidate pairs from the object detector.

\section{Ablation studies}
\label{part:supmat_ablation}

In this section, we perform ablation studies that complement the analysis in Section~\ref{part:experiments} of the main paper. We discuss the benefits of the different components of our model introduced in Section~\ref{part:model_1} of the main paper, and in particular the benefits of the visual phrase module. We also analyze the influence of pre-trained word embeddings and the effect of adding bigrams modules.

\begin{table}[t]
\centering
\small{
\begin{tabular}{@{}lrrrr@{}}\toprule
											& & full 	& rare 	& non-rare \\\midrule
(a) & \text{s+o (obj.det.)}						&  5.6 & 4.2 & 6.5 	\\
(b) & \text{s+o}									& 10.0	& 7.6	& 10.8	\\
(c) & \text{p}									& 14.9 & 9.4  & 16.5 \\
(d) & \text{bigrams}								& 14.9 & 9.6  & 16.5 \\
(e) & \text{vp}									& 16.5	& 10.4	& 18.4	\\
(f) & \text{s+o+vp (\textit{main paper})}  				& 17.7 & 11.6	& 19.5 \\
(g) & \text{s+o+p (classifier)}					& 18.0 & 13.4 & 19.4  	\\
(h) & \text{s+o+p (random words)}				& 18.4 & 13.7 & 19.8 \\
(i) & \text{s+o+p (\textit{main paper})}					& 18.7 & 13.8	& 20.1 \\
(j) & \text{s+o+p (finetuned words)}				& 18.8 & 14.5	& 20.1 \\
(k) & \text{s+o+p+vp (\textit{main paper})}				&  19.4   & 14.6    & 20.9 	\\
(l) & \text{s+o+p+bigrams}						&  19.5   & 14.6    & 21.0 	\\
(m) & \text{s+o+p+vp+bigrams}					&  \textbf{20.0}  & \textbf{15.0}  & \textbf{21.5} 	\\
\bottomrule
\end{tabular}
}
\vspace{-1.5ex}
\caption{\small Ablation study on HICO-DET.}
\vspace{-4ex}
\label{tab:ablation studies}
\end{table}

\vspace{-0.3cm}
\paragraph{Benefits of different components of our model.}
Our contribution is a hybrid model which combines subject (\textit{s}), object (\textit{o}), predicate (\textit{p}) and visual phrase (\textit{vp}) modules. We show in Table~\ref{tab:ablation studies}, which complements Table 1 of the main paper, that each of these modules is making a complementary contribution. 
The performance of our compositional model \textit{s+o+p} builds on our strong unigram models \textit{s+o} (row (b)) that already significantly improve over the baseline using only the object scores returned by pre-trained object detectors (row (a)) typically used by other methods~\cite{iCAN,Gkioxari18}. 
The strength of our modules for representing visual relations is clearly demonstrated by the good performance of our unigram predicate model \textit{p} (row (c)) and the visual phrase model \textit{vp} (row (e)) over using objects alone (cf. \textit{s+o}, row (b)). In addition, \textit{vp} alone performs better than \textit{p} alone (row (e) $>$ (c)). 
Importantly, these modules are complementary as clearly shown by the best performance of our combined model (row (k)) that can also easily incorporate bigrams (row (m)), see below. 

\begin{figure}[t]
\centering
    	\vspace{-1.5ex}        
    \begin{minipage}[t]{0.15\textwidth}
    	\centering
    	\includegraphics[height=2.2cm]{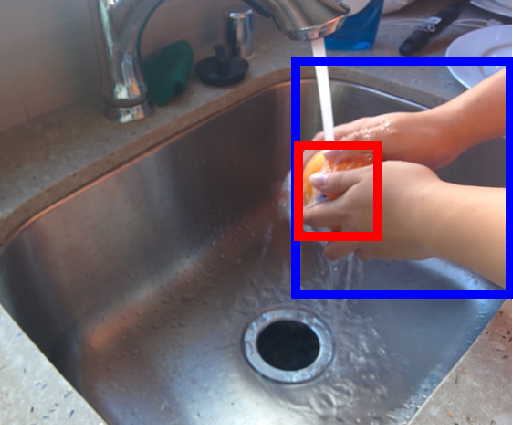}\\	
    	\footnotesize{(a) \color{blue}{person} \textcolor{Green}{wash} \color{red}{orange}}
    \end{minipage}
    \begin{minipage}[t]{0.15\textwidth}
    	\centering
    	\includegraphics[height=2.2cm]{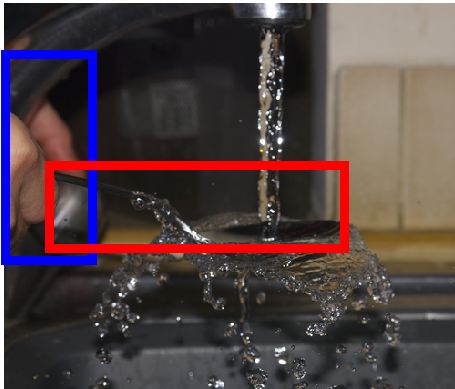}\\
    	\footnotesize{(b) \color{blue}{person} \textcolor{Green}{wash} \color{red}{spoon}}
    \end{minipage} 
    \begin{minipage}[t]{0.16\textwidth}
    	\centering
    	\includegraphics[height=2.2cm]{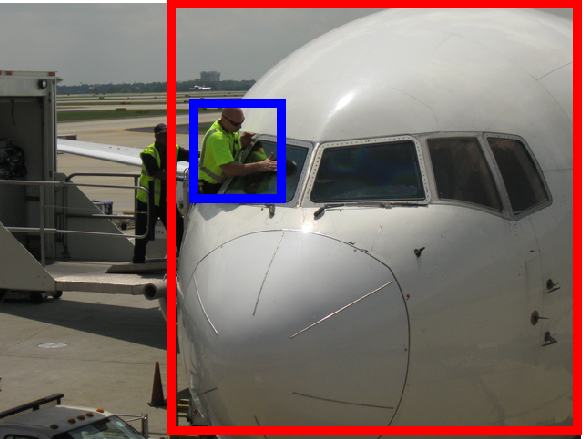}\\	
    	\footnotesize{(c) \color{blue}{person} \textcolor{Green}{wash} \color{red}{airplane}}
    \end{minipage}
    \vspace{-1ex}
    \caption{\small Retrieval examples where \textit{s+o+p+vp} is better than \textit{s+o+p}.}
    \label{fig:vp}
    \vspace{-2ex}
\end{figure}

\vspace{-0.3cm}
\paragraph{Benefits of visual phrase (vp) model.}
The improvement thanks to the \textit{vp} model is consistent over several datasets. We found qualitatively that the \textit{vp} branch handles important unusual situations where the compositional model (\textit{s+o+p}) fails, which happens when (at least) one of the \textit{s}, \textit{o} and \textit{p} branches has a low score, e.g. due to object occlusion (Figure~\ref{fig:vp}(a)), unusual object appearance (Figure~\ref{fig:vp}(b)) or unusual spatial configuration (Figure~\ref{fig:vp}(c)). The visual phrase model (\textit{vp}) can better handle these situations because it better models the specific appearance and spatial configuration of triplets seen in training.

\vspace{-0.3cm}
\paragraph{Benefits of word vectors.}
In Table~\ref{tab:ablation studies} we (1) show benefits of mapping input triplets to image-language embedding space instead of learning {\it s}, {\it o} and {\it p} classifiers (row (h) $>$ (g)) and (2) confirm that using pre-trained word embeddings helps, but only slightly (row (i) $>$ (h)).
Because of the mismatch between word usage in the pre-training text corpus and our dataset, we also found that fine-tuning the pre-trained word embeddings is beneficial (row (j) $>$ row (i)). 

\vspace{-0.3cm}
\paragraph{Incorporating bigrams.}
While our primary focus is to marry compositional (unigrams) and visual phrase (trigram) models, we can easily incorporate bigram branches (\textit{sp+po}) in our model. 
As shown in Table~\ref{tab:ablation studies}, bigrams provide an improvement over unigrams modelling subject and object independently \textit{s+o} (row (d) $>$ (b)) and combined with unigrams (row (l)) they reach comparable results to a combination of unigrams and trigram (row (k)). Interestingly, bigrams and trigram are complementary. Their combination leads to the overall best results (row (m)).

\vspace{-0.3cm}
\paragraph{Alternative to weighting function G.}
We tested an alternative to the weighting function G (Eq.~\eqref{eq:similarity} of the main paper) taking as input word2vec embeddings instead of joint visual-semantic embeddings in $s$, $o$ and $p$ spaces. This lead to a slight performance drop (28.3 vs. 28.6 for our analogy transfer in Table~\ref{tab:results_hico_zeroshot} of the main paper). This result suggests that while pre-trained language embeddings are core ingredients to establish similarities between concepts, they can be further strengthen by using visual appearance.

\section{Qualitative results on HICO-DET dataset}
\label{part:supmat_hico}

In Figure~\ref{fig:qualitative_hico_supmat} we show additional examples of retrieved detections for unseen triplets that supplement Figure~\ref{fig:qualitative_hico} of the main paper. These qualitative examples confirm that our model for transfer by analogy (s+o+vp+transfer) (Section~\ref{part:model_2} of the main paper) automatically selects relevant source triplets (S) given an unseen triplet query (Q). For instance, for the query triplet ``person throw frisbee" (first row), our model selects (1) a source triplet that involves the same action, with a different, but similar, object ``person throw sports ball", (2) two source triplets with the same object, and different, but related, actions ``person catch frisbee", ``person block frisbee" and (3) two other source triplets with different, but related, object and actions ``person hit sports ball", ``person serve sports ball". Similar conclusions hold for the other examples displayed. The top false positives indicate that the main failure mode is the confusion with another similar interaction (e.g. ``lie on" is confused with ``sit on" in row 3 or ``inspect" is confused with ``hold" in row 4. Some detections are also incorrectly classified as failure, as they are still some missing ground truth annotations (e.g. row 2, row 6).

\section{Qualitative results on UnRel dataset}
\label{part:supmat_unrel}

In Figure~\ref{fig:qualitative_unrel_supmat} we show additional qualitative results for our model (p+vp+transfer) for retrieval of unseen (unusual) triplets on the UnRel dataset supplementing results shown in Figure~\ref{fig:qualitative_unrel} of the main paper. We show the source triplets (S) automatically sampled by our analogy model that are used to form the visual phrase embedding of the target query (Q). The top true positive retrievals are shown in green, the top false positive retrieval is shown in red. 
The automatically sampled source triplets all appear relevant. Our method samples source triplets involving (1) a different subject (``dog ride bike" is transferred from ``person ride bike", ``building has wheel" is transferred from ``truck has wheel"), (2) a different object (``person stand on horse" is transferred from ``person stand on sand"), or (3) a different predicate (``cone on the top of person" is transferred from ``sky over person"). The results confirm that our model works well not only for human-object interactions but also for more general interactions involving spatial relations (e.g. ``in", ``on the top of") or a subject different from a person (e.g. ``cone", ``car", ``building", ``dog"). 
There are two main failure modes illustrated by the top false positive detections. The first one is an incorrect object detection (e.g. ``train" is confused with ``building" in row 3, or ``motorcycle" is confused with ``bike" in row 2).  
The second failure mode is due to the confusion with another similar triplet, possibly due to the unusual character of UnRel queries which sometimes make it difficult to sample close enough source triplets for the transfer by analogy. For instance, it is hard to form a good embedding for ``car in building" from source triplets ``car in street", ``bus in street", ``person in street" as these source triplets have fairly different visual appearance (row 5).

\section{Qualitative results on COCO-a dataset}
\label{part:supmat_cocoa}

In Figure~\ref{fig:qualitative_cocoa_supmat}, we show additional qualitative results of our model for transfer by analogy (s+o+vp+transfer) on retrieval of unseen (out of vocabulary) triplets in the COCO-a dataset, complementing results in Figure~\ref{fig:qualitative_cocoa} of the main paper. We display the source triplets (S) automatically sampled by our model for a target query (Q). Despite the fact that the target predicates are not seen in training, our model manages, most of the time, to sample relevant source triplets for transfer. For instance, our model would link the unseen triplet ``person use laptop", involving the unseen predicate ``use" (row 2) to source triplets such as ``person type on laptop", ``person read laptop" or ``person text on phone", all involving a predicate that is relevant to the unseen target predicate ``use''. The same holds for the unseen triplet ``person touch horse" (row 3) for which our model samples source triplets involving contact interaction such as ``person hug horse", ``person pet horse" or ``person kiss horse". The top false detections are informative : (i) either they correspond to interactions involving related triplets, which are likely to be sampled as source triplets (e.g. ``person shear sheep" confused with ``person caress sheep" in row 1), (ii) or they correspond to interactions with ambiguous semantics (e.g. ``person get frisbee" or ``person prepare kite" that involve ambiguous predicates that could correspond to a large variety of spatial configurations).

\section{Visualization of joint embedding spaces}
\label{part:supmat_tsne}

\begin{figure*}[t]
\includegraphics[trim={0.5cm 0cm 7cm 0cm},clip,width=\linewidth]{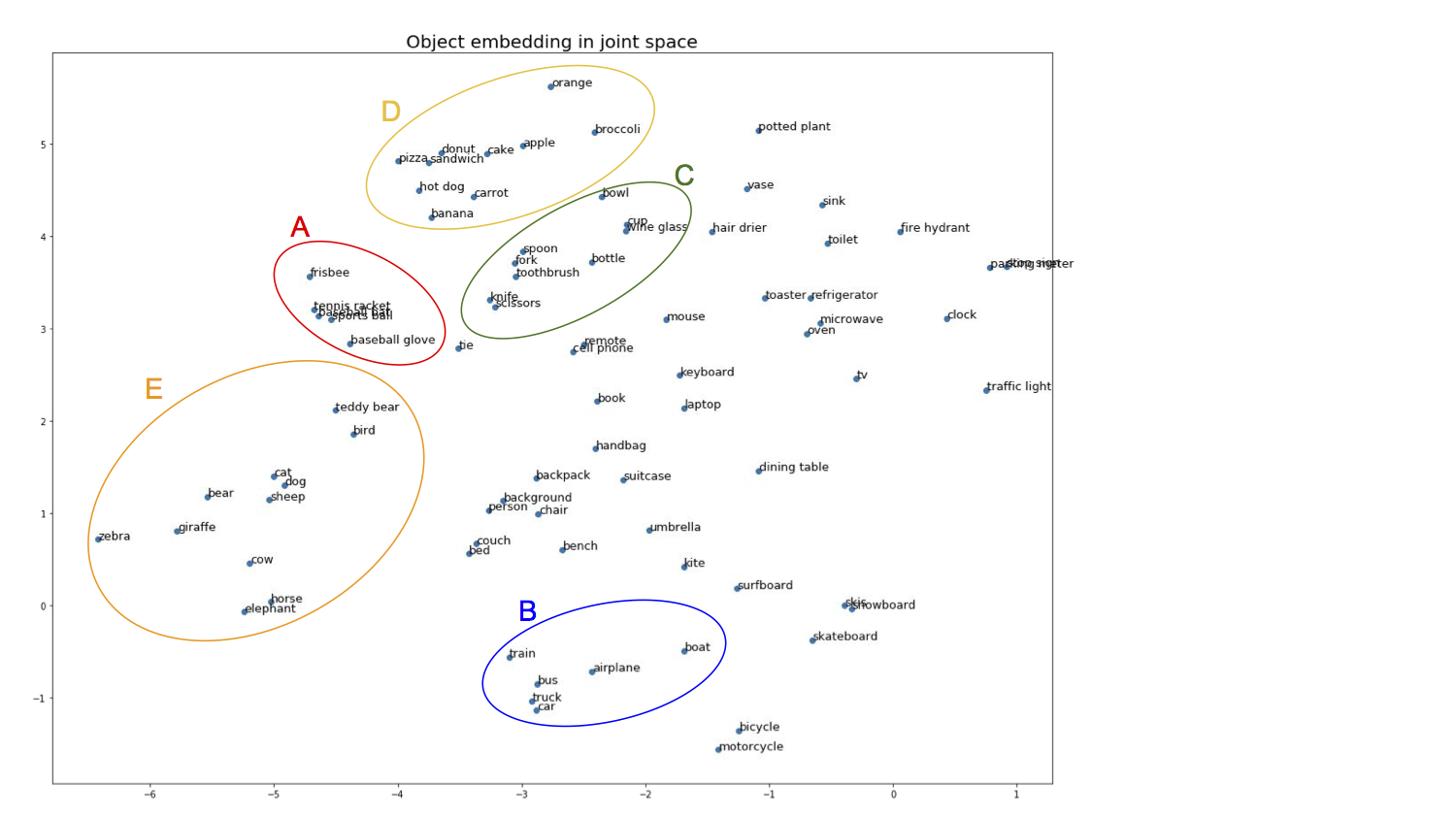}\\\caption{Object embedding on HICO-DET visualized using T-sne~\cite{tsne}. Objects appear to be grouped according to their visual and semantic similarity. For example, we highlight regions corresponding to: (A) sports instruments (e.g. ``tennis racket", ``frisbee"), (B) big transportation (e.g. ``bus", ``train"), (C) eating utensils (e.g. ``fork", ``cup"), (D) food (e.g. ``pizza", apple"), (E) animals (e.g. ``giraffe", ``bird"). Learning a good embedding for unigrams (here objects) is crucial in our model that uses the transfer by analogy, as unigram embeddings directly influence the analogy transformation from the seen visual phrases to the unseen ones.}
\label{fig:tsne_o}
\end{figure*}

\begin{figure*}[t]
\includegraphics[trim={0.5cm 0cm 7cm 0cm},clip,width=\linewidth]{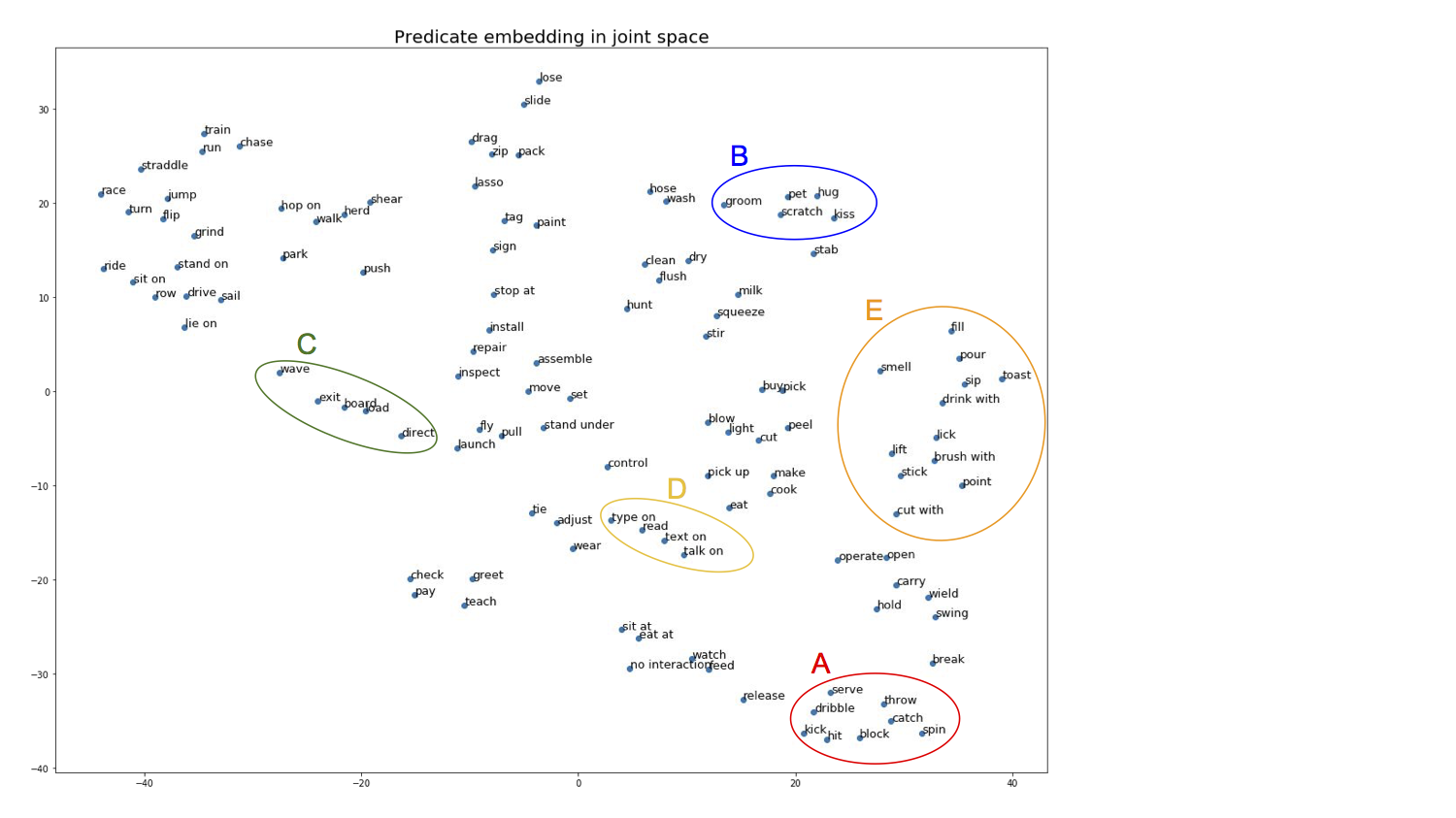}\\\caption{Predicate embedding on HICO-DET visualized with T-sne~\cite{tsne}. The predicates are grouped according to their visual and semantic similarity. For example, we highlight regions corresponding to: (A) interactions related to sports (e.g. ``throw", ``dribble"), (B) gentle interactions with an animal/person (e.g. ``hug", ``kiss"), (C) interactions with transportation vehicles (e.g. ``board", ``exit"), (D) interactions with (electronic) devices (e.g. ``text on", ``read"), (E) interactions with food (e.g. ``smell", ``lick"). Learning a good embedding for unigrams (here predicates) is crucial in our model that uses transfer by analogy, as unigram embeddings directly influence the analogy transformation from the seen visual phrases to the unseen ones.}
\label{fig:tsne_p}
\end{figure*}

\begin{figure*}[t]
\includegraphics[trim={0cm 0cm 0.1cm 0cm},clip,width=\linewidth]{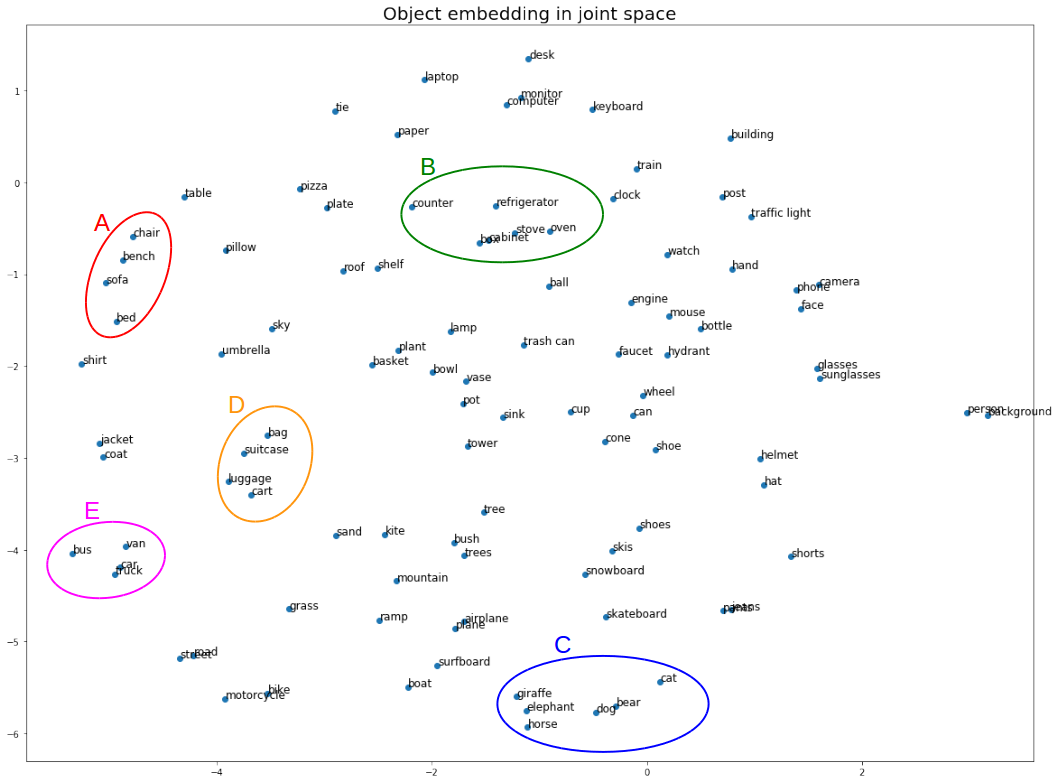}\\\caption{Object embedding on the UnRel dataset visualized using T-sne~\cite{tsne}. Objects appear to be grouped according to their visual and semantic similarity. For example, we highlight regions corresponding to: (A) piece of furniture on which to sit (e.g. ``chair", ``bench"), (B) kitchen furniture (e.g. ``refrigerator", ``stove"), (C) animals (e.g. ``giraffe", ``cat"), (D) bags and containers (e.g. ``suitcase", cart"), (E) motorized transportation (e.g. ``bus", ``car"). Learning a good embedding for unigrams (here objects) is crucial in our model that uses the transfer by analogy, as unigram embeddings directly influence the analogy transformation from the seen visual phrases to the unseen ones.}
\label{fig:tsne_o_vrd}
\end{figure*}

\begin{figure*}[t]
\includegraphics[trim={0cm 0cm 0.1cm 0cm},clip,width=\linewidth]{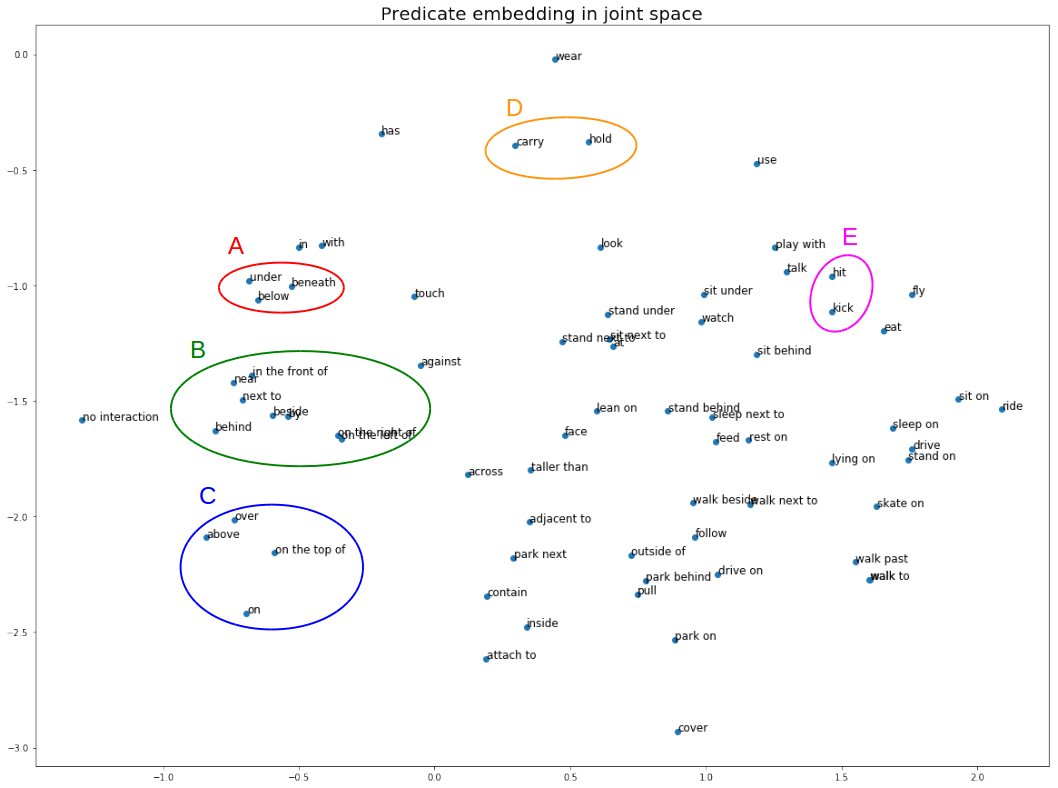}\\\caption{Predicate embedding on the UnRel dataset visualized with T-sne~\cite{tsne}. The predicates are grouped according to their visual and semantic similarity. For example, we highlight regions corresponding to: (A) spatial relations related to ``under" (e.g. ``below", ``beneath"), (B) spatial relations related to ``next to" (e.g. ``near", ``beside"), (C) spatial relations related to ``above" (e.g. ``over", ``on the top of"), or similar actions (D) and (E). Note that it is remarkable that our visual-semantic embedding separates relations such as those in (A) from those in (C) while they are very similar from a strictly semantic point of view (in pre-trained word2vec embeddings). Learning a good embedding for unigrams (here predicates) is crucial in our model that uses transfer by analogy, as unigram embeddings directly influence the analogy transformation from the seen visual phrases to the unseen ones.}
\label{fig:tsne_p_vrd}
\end{figure*}

Here, we provide additional insights about the embedding spaces learnt on the HICO-DET dataset and UnRel dataset using the t-sne visualization~\cite{tsne} of the final learnt joint embedding. 
First, we show t-sne visualization~\cite{tsne} of joint embedding spaces learnt for objects and predicates on HICO-DET to better understand which concepts are close together in the learnt space. For the object embedding, as shown in Figure~\ref{fig:tsne_o}, objects are grouped according to their visual and semantic similarity. The same holds for predicate embeddings shown in Figure~\ref{fig:tsne_p}. We draw similar plots for UnRel dataset, showing the object embedding in Figure~\ref{fig:tsne_o_vrd} and the predicate embedding in Figure~\ref{fig:tsne_p_vrd}. The visualization of predicate embedding on UnRel dataset in Figure~\ref{fig:tsne_p_vrd} is especially interesting as it involves spatial relations. We remark that our model is able to separate spatial relations such as ``under" from ``above" which are semantically very similar. Learning good embedding for unigrams is crucial in our model for transfer by analogy, as unigram embeddings directly influence the analogy transformation from the seen visual phrases to the unseen ones.

\input{supmat_fig_hico}
\input{supmat_fig_unrel}
\input{supmat_fig_cocoa}

%% file: supmat_fig_hico.tex
\begin{figure*}[t]
\centering
	\begin{minipage}[t]{0.22\textwidth}
    \centering
    	\textit{Query (Q) / Source (S)}\\
    	\vspace{2ex}
	\end{minipage}	
	\hspace{0.01\textwidth}
	\begin{minipage}[t]{0.56\textwidth}
    	\centering
    \textit{	Top true positives}\\
    	\vspace{2ex}
	\end{minipage}
	\hspace{0.005\textwidth}
	\begin{minipage}[t]{0.18\textwidth}
    \centering
    \textit{	Top false positive}\\
    	\vspace{2ex}
	\end{minipage}

    \begin{minipage}[c]{0.24\textwidth}
    \vspace{-12ex}
    \small{
	(Q) \textbf{{\color{blue}person} {\textcolor{Green}{throw}} {\color{red}frisbee}} \\     	
	\vspace{-18pt}
	\par\noindent\rule{\textwidth}{0.4pt}
	(S) {\color{blue}person} {\textcolor{Green}{throw}} {\color{red}sports ball} \\
	(S) {\color{blue}person} {\textcolor{Green}{catch}} {\color{red}frisbee} \\
	(S) {\color{blue}person} {\textcolor{Green}{block}} {\color{red}frisbee} \\
	(S) {\color{blue}person} {\textcolor{Green}{hit}} {\color{red}sports ball} \\
	(S) {\color{blue}person} {\textcolor{Green}{serve}} {\color{red}sports ball} \\
	}
    \end{minipage}  
    \begin{minipage}[t]{0.18\textwidth}
    	\centering
       	\includegraphics[trim={1.1cm 0.6cm 0.75cm 0.5cm},clip,width=0.95\linewidth,cfbox={green 2pt 2pt}]{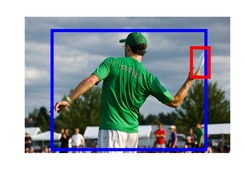}\\
       	\vspace{1.5ex}
    \end{minipage}
    \hspace{0.0025\textwidth}
    \begin{minipage}[t]{0.18\textwidth}
    	\centering
       	\includegraphics[trim={0.95cm 0.5cm 0.6cm 0.4cm},clip,width=0.95\linewidth,cfbox={green 2pt 2pt}]{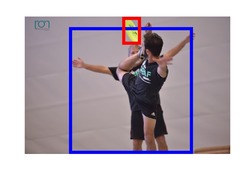}\\
       	\vspace{1.5ex}
    \end{minipage}
    \hspace{0.0025\textwidth}
    \begin{minipage}[t]{0.18\textwidth}
       \centering
       \includegraphics[trim={1.5cm 0.3cm 0.5cm 0.1cm},clip,width=0.95\linewidth,cfbox={green 2pt 2pt}]{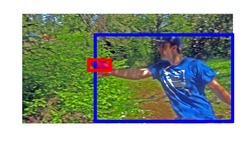}\\
       \vspace{1.5ex}
    \end{minipage}
    \hspace{0.0025\textwidth}
    \begin{minipage}[t]{0.18\textwidth}
    	\centering
       	\includegraphics[trim={-0.75cm 0.8cm -0.75cm 0.6cm},clip,width=0.95\linewidth,cfbox={red 2pt 2pt}]{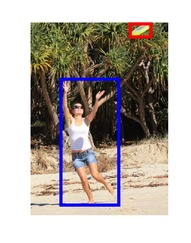}\\
      	\vspace{1.5ex}
    \end{minipage}

    \begin{minipage}[c]{0.24\textwidth}
    \vspace{-10ex}	
    		\small{
	(Q) \textbf{{\color{blue}person} {\textcolor{Green}{hold}} {\color{red}surfboard}}\\
	\vspace{-18pt}
	\par\noindent\rule{\textwidth}{0.4pt}
	(S) {\color{blue}person} {\textcolor{Green}{hold}} {\color{red}frisbee} \\
	(S) {\color{blue}person} {\textcolor{Green}{hold}} {\color{red}kite} \\
	(S) {\color{blue}person} {\textcolor{Green}{hold}} {\color{red}umbrella} \\
	(S) {\color{blue}person} {\textcolor{Green}{hold}} {\color{red}snowboard} \\
	(S) {\color{blue}person} {\textcolor{Green}{hold}} {\color{red}skis} \\
		}		
    \end{minipage}  
    \begin{minipage}[t]{0.18\textwidth}
    	\centering
       	\includegraphics[trim={0.75cm 0.5cm 0.75cm 0.6cm},clip,width=0.95\linewidth,cfbox={green 2pt 2pt}]{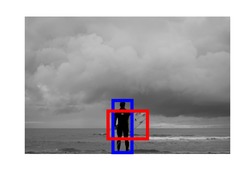}\\
       	\vspace{1.5ex}
    \end{minipage}
    \hspace{0.0025\textwidth}
    \begin{minipage}[t]{0.18\textwidth}
    	\centering
       	\includegraphics[trim={0.7cm 0.5cm 0.65cm 0.5cmm},clip,width=0.95\linewidth,cfbox={green 2pt 2pt}]{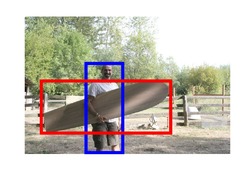}\\
       	\vspace{1.5ex}
    \end{minipage}
    \hspace{0.0025\textwidth}
    \begin{minipage}[t]{0.18\textwidth}
       \centering
       \includegraphics[trim={1.37cm 1cm 0.5cm 0.5cm},clip,width=0.95\linewidth,cfbox={green 2pt 2pt}]{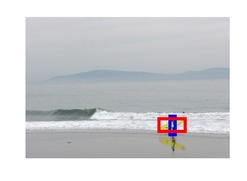}\\
       \vspace{1.5ex}
    \end{minipage}
    \hspace{0.0025\textwidth}
    \begin{minipage}[t]{0.18\textwidth}
    	\centering
       	\includegraphics[trim={0.8cm 1cm 0.5cm 0.9cm},clip,width=0.95\linewidth,cfbox={red 2pt 2pt}]{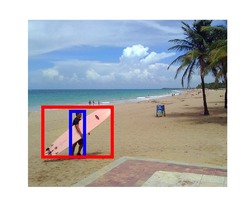}\\
      	\vspace{1.5ex}
    \end{minipage}

    \begin{minipage}[c]{0.24\textwidth}
    \vspace{-12ex}	
    		\small{
	(Q) \textbf{{\color{blue}person} {\textcolor{Green}{lie on}} {\color{red}bed}}\\
	\vspace{-18pt}
	\par\noindent\rule{\textwidth}{0.4pt}
	(S) {\color{blue}person} {\textcolor{Green}{lie on}} {\color{red}couch} \\
	(S) {\color{blue}person} {\textcolor{Green}{lie on}} {\color{red}chair} \\
	(S) {\color{blue}person} {\textcolor{Green}{lie on}} {\color{red}bench} \\
	(S) {\color{blue}person} {\textcolor{Green}{lie on}} {\color{red}surfboard} \\
	(S) {\color{blue}person} {\textcolor{Green}{sit on}} {\color{red}bed} \\
		}		
    \end{minipage}  
    \begin{minipage}[t]{0.18\textwidth}
    	\centering
       	\includegraphics[trim={0.7cm 0.6cm 0.5cm 0.5cm},clip,width=0.95\linewidth,cfbox={green 2pt 2pt}]{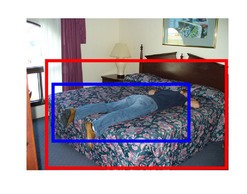}\\
       	\vspace{1.5ex}
    \end{minipage}
    \hspace{0.0025\textwidth}
    \begin{minipage}[t]{0.18\textwidth}
    	\centering
       	\includegraphics[trim={0cm 0.7cm 0cm 1.7cm},clip,width=0.95\linewidth,cfbox={green 2pt 2pt}]{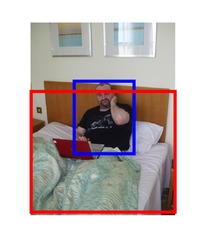}\\
       	\vspace{1.5ex}
    \end{minipage}
    \hspace{0.0025\textwidth}
    \begin{minipage}[t]{0.18\textwidth}
       \centering
       \includegraphics[trim={0.65cm 0.6cm 0.55cm 0.5cm},clip,width=0.95\linewidth,cfbox={green 2pt 2pt}]{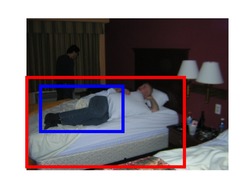}\\
       \vspace{1.5ex}
    \end{minipage}
    \hspace{0.0025\textwidth}
    \begin{minipage}[t]{0.18\textwidth}
    	\centering
       	\includegraphics[trim={1.4cm 0.6cm 0.4cm 0.5cm},clip,width=0.95\linewidth,cfbox={red 2pt 2pt}]{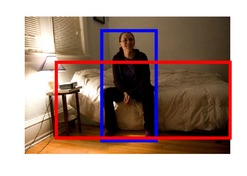}\\
      	\vspace{1.5ex}
    \end{minipage}  
 
    \begin{minipage}[c]{0.24\textwidth}
    \vspace{-12ex}	
    		\small{
	(Q) \textbf{{\color{blue}person} {\textcolor{Green}{inspect}} {\color{red}bicycle}}\\
	\vspace{-18pt}
	\par\noindent\rule{\textwidth}{0.4pt}
	(S) {\color{blue}person} {\textcolor{Green}{inspect}} {\color{red}motorcycle} \\
	(S) {\color{blue}person} {\textcolor{Green}{inspect}} {\color{red}bus} \\
	(S) {\color{blue}person} {\textcolor{Green}{inspect}} {\color{red}dog} \\
	(S) {\color{blue}person} {\textcolor{Green}{inspect}} {\color{red}backpack} \\
	(S) {\color{blue}person} {\textcolor{Green}{inspect}} {\color{red}car} \\
		}		
    \end{minipage}  
    \begin{minipage}[t]{0.18\textwidth}
    	\centering
       	\includegraphics[trim={1cm 0.6cm 0.5cm 0.6cm},clip,width=0.95\linewidth,cfbox={green 2pt 2pt}]{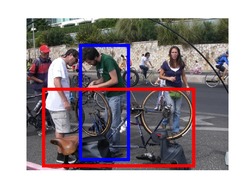}\\
       	\vspace{1.5ex}
    \end{minipage}
    \hspace{0.0025\textwidth}
    \begin{minipage}[t]{0.18\textwidth}
    	\centering
       	\includegraphics[trim={0cm 1.5cm 0cm 1.5cmm},clip,width=0.95\linewidth,cfbox={green 2pt 2pt}]{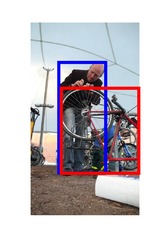}\\
       	\vspace{1.5ex}
    \end{minipage}
    \hspace{0.0025\textwidth}
    \begin{minipage}[t]{0.18\textwidth}
       \centering
       \includegraphics[trim={0.7cm 1cm 0.5cm 2.22cm},clip,width=0.95\linewidth,cfbox={green 2pt 2pt}]{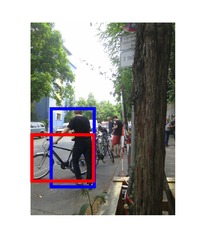}\\
       \vspace{1.5ex}
    \end{minipage}
    \hspace{0.0025\textwidth}
    \begin{minipage}[t]{0.18\textwidth}
    	\centering
       	\includegraphics[trim={-0.2cm 1cm -0.3cm 1.2cm},clip,width=0.95\linewidth,cfbox={red 2pt 2pt}]{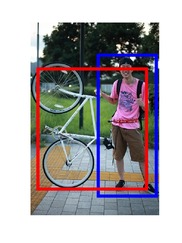}\\
      	\vspace{1.5ex}
    \end{minipage}

    \begin{minipage}[c]{0.24\textwidth}
    \vspace{-13ex}	
    		\small{
	(Q) \textbf{{\color{blue}person} {\textcolor{Green}{hug}} {\color{red}dog}}\\
	\vspace{-18pt}
	\par\noindent\rule{\textwidth}{0.4pt}
	(S) {\color{blue}person} {\textcolor{Green}{hug}} {\color{red}cat} \\
	(S) {\color{blue}person} {\textcolor{Green}{hug}} {\color{red}sheep} \\
	(S) {\color{blue}person} {\textcolor{Green}{hug}} {\color{red}teddy bear} \\
	(S) {\color{blue}person} {\textcolor{Green}{hug}} {\color{red}horse} \\
	(S) {\color{blue}person} {\textcolor{Green}{hug}} {\color{red}person} \\
		}		
    \end{minipage}  
    \begin{minipage}[t]{0.18\textwidth}
    	\centering
       	\includegraphics[trim={-.50cm 0.7cm -.50cm 0.7cm},clip,width=0.95\linewidth,cfbox={green 2pt 2pt}]{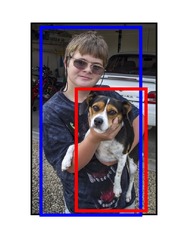}\\
       	\vspace{1.5ex}
    \end{minipage}
    \hspace{0.0025\textwidth}
    \begin{minipage}[t]{0.18\textwidth}
    	\centering
       	\includegraphics[trim={-0.3cm 0.7cm -0.35cm 0.6cm},clip,width=0.95\linewidth,cfbox={green 2pt 2pt}]{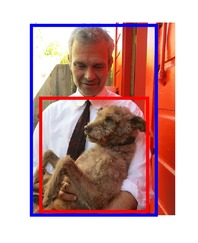}\\
       	\vspace{1.5ex}
    \end{minipage}
    \hspace{0.0025\textwidth}
    \begin{minipage}[t]{0.18\textwidth}
       \centering
       \includegraphics[trim={1.15cm 0.55cm 0.5cm 0.4cm},clip,width=0.95\linewidth,cfbox={green 2pt 2pt}]{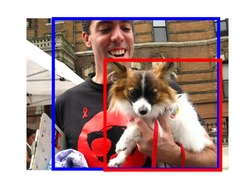}\\
       \vspace{1.5ex}
    \end{minipage}
    \hspace{0.0025\textwidth}
    \begin{minipage}[t]{0.18\textwidth}
    	\centering
       	\includegraphics[trim={1.35cm 0.6cm 0.45cm 0.45cm},clip,width=0.95\linewidth,cfbox={red 2pt 2pt}]{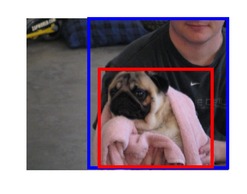}\\
      	\vspace{1.5ex}
    \end{minipage}

    \begin{minipage}[c]{0.24\textwidth}
    \vspace{-12ex}	
    		\small{
	(Q) \textbf{{\color{blue}person} {\textcolor{Green}{straddle}} {\color{red}motorcycle}}\\
	\vspace{-18pt}
	\par\noindent\rule{\textwidth}{0.4pt}
	(S) {\color{blue}person} {\textcolor{Green}{straddle}} {\color{red}horse} \\
	(S) {\color{blue}person} {\textcolor{Green}{straddle}} {\color{red}bicycle} \\
	(S) {\color{blue}person} {\textcolor{Green}{straddle}} {\color{red}dog} \\
	(S) {\color{blue}person} {\textcolor{Green}{push}} {\color{red}motorcycle} \\
	(S) {\color{blue}person} {\textcolor{Green}{turn}} {\color{red}motorcycle} \\
		}		
    \end{minipage}  
    \begin{minipage}[t]{0.18\textwidth}
    	\centering
       	\includegraphics[trim={0cm 0.8cm 0cm 1.6cm},clip,width=0.95\linewidth,cfbox={green 2pt 2pt}]{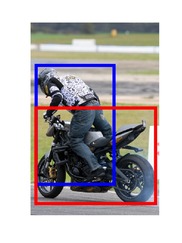}\\
       	\vspace{1.5ex}
    \end{minipage}
    \hspace{0.0025\textwidth}
    \begin{minipage}[t]{0.18\textwidth}
    	\centering
       	\includegraphics[trim={1.3cm 0.7cm 1cm 0.5cm},clip,width=0.95\linewidth,cfbox={green 2pt 2pt}]{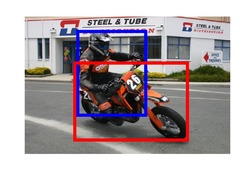}\\
       	\vspace{1.5ex}
    \end{minipage}
    \hspace{0.0025\textwidth}
    \begin{minipage}[t]{0.18\textwidth}
       \centering
       \includegraphics[trim={0.1cm 1cm 0cm 1.38cm},clip,width=0.95\linewidth,cfbox={green 2pt 2pt}]{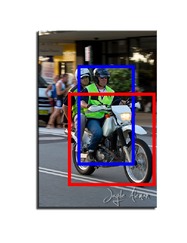}\\
       \vspace{1.5ex}
    \end{minipage}
    \hspace{0.0025\textwidth}
    \begin{minipage}[t]{0.18\textwidth}
    	\centering
       	\includegraphics[trim={1cm 0.7cm 1.05cm 0.5cm},clip,width=0.95\linewidth,cfbox={red 2pt 2pt}]{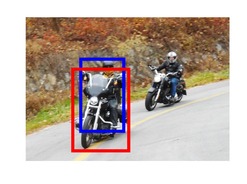}\\
      	\vspace{1.5ex}
    \end{minipage}

    \setlength\abovecaptionskip{10pt}
    \caption{\small {\bf Retrieval examples on the HICO-DET dataset.} Top retrieved positives (green) and negatives (red) using our model (s+o+vp+transfer) for unseen triplet queries. The query is marked as (Q). The source triplets automatically selected by our model are marked as (S).
  For instance, for the query triplet ``person throw frisbee" (first row), our model selects (1) a source triplet that involves the same action, with a different, but similar, object ``person throw sports ball", (2) two source triplets with the same object, and different, but related, actions ``person catch frisbee", ``person block frisbee" and (3) two other source triplets with different, but related, object and actions ``person hit sports ball", ``person serve sports ball". The top false positives show the main failure mode: the interaction is confused with another similar interaction (e.g. ``lie on" is confused with ``sit on" in row 3 or ``inspect" is confused with ``hold" in row 4). Also, we note that some mistakes among the top false positives are due to missing ground truth annotations.}
    \label{fig:qualitative_hico_supmat}
\end{figure*}

%% file: supmat_fig_unrel.tex
\begin{figure*}[t]
\centering
	\begin{minipage}[t]{0.22\textwidth}
    \centering
    	\textit{Query (Q) / Source (S)}\\
    	\vspace{2ex}
	\end{minipage}	
	\hspace{0.01\textwidth}
	\begin{minipage}[t]{0.56\textwidth}
    	\centering
    \textit{	Top true positives}\\
    	\vspace{2ex}
	\end{minipage}
	\hspace{0.005\textwidth}
	\begin{minipage}[t]{0.18\textwidth}
    \centering
    \textit{	Top false positive}\\
    	\vspace{2ex}
	\end{minipage}

    \begin{minipage}[c]{0.24\textwidth}
    \vspace{-10ex}
    \small{
	(Q) \textbf{{\color{blue}person} {\textcolor{Green}{stand on}} {\color{red}horse}} \\     	
	\vspace{-18pt}
	\par\noindent\rule{\textwidth}{0.4pt}
	(S) {\color{blue}person} {\textcolor{Green}{stand on}} {\color{red}sand} \\
	(S) {\color{blue}person} {\textcolor{Green}{stand on}} {\color{red}grass} \\
	(S) {\color{blue}person} {\textcolor{Green}{stand on}} {\color{red}street} \\
	(S) {\color{blue}person} {\textcolor{Green}{sit on}} {\color{red}motorcycle} \\
	(S) {\color{blue}person} {\textcolor{Green}{sit on}} {\color{red}bench} \\
	}
    \end{minipage}  
    \begin{minipage}[t]{0.18\textwidth}
    	\centering
       	\includegraphics[trim={1.2cm 0.8cm 1.8cm 0.5cm},clip,width=0.95\linewidth,cfbox={green 2pt 2pt}]{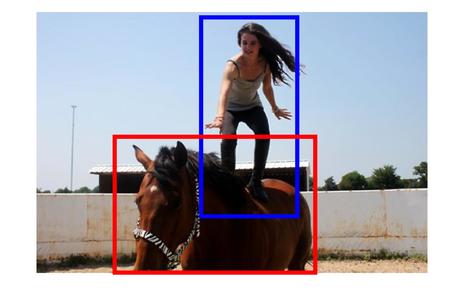}\\
       	\vspace{1.5ex}
    \end{minipage}
    \hspace{0.0025\textwidth}
    \begin{minipage}[t]{0.18\textwidth}
    	\centering
       	\includegraphics[trim={1.2cm 1.1cm 1.2cm 0.9cm},clip,width=0.95\linewidth,cfbox={green 2pt 2pt}]{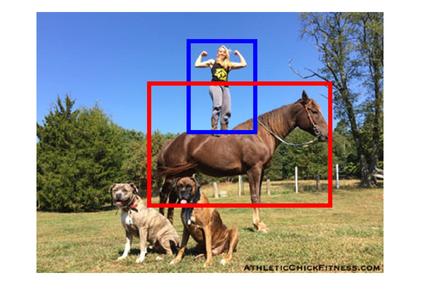}\\
       	\vspace{1.5ex}
    \end{minipage}
    \hspace{0.0025\textwidth}
    \begin{minipage}[t]{0.18\textwidth}
       \centering
       \includegraphics[trim={1.5cm 0.8cm 1.4cm 0.5cm},clip,width=0.95\linewidth,cfbox={green 2pt 2pt}]{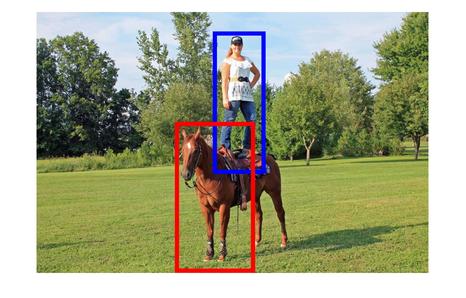}\\
       \vspace{1.5ex}
    \end{minipage}
    \hspace{0.0025\textwidth}
    \begin{minipage}[t]{0.18\textwidth}
    	\centering
       	\includegraphics[trim={1.5cm 0.8cm 1.5cm 0.6cm},clip,width=0.95\linewidth,cfbox={red 2pt 2pt}]{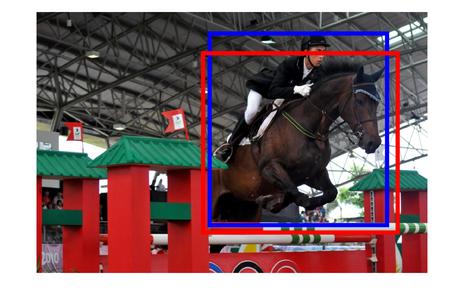}\\
      	\vspace{1.5ex}
    \end{minipage}

    \begin{minipage}[c]{0.24\textwidth}
    \vspace{-13ex}
    \small{
	(Q) \textbf{{\color{blue}dog} {\textcolor{Green}{ride}} {\color{red}bike}} \\     	
	\vspace{-18pt}
	\par\noindent\rule{\textwidth}{0.4pt}
	(S) {\color{blue}person} {\textcolor{Green}{ride}} {\color{red}bike} \\
	(S) {\color{blue}person} {\textcolor{Green}{ride}} {\color{red}motorcycle} \\
	(S) {\color{blue}person} {\textcolor{Green}{ride}} {\color{red}skateboard} \\
	(S) {\color{blue}person} {\textcolor{Green}{ride}} {\color{red}horse} \\
	(S) {\color{blue}person} {\textcolor{Green}{ride}} {\color{red}boat} \\
	}
    \end{minipage}  
    \begin{minipage}[t]{0.18\textwidth}
    	\centering
       	\includegraphics[trim={0cm 2cm 0cm 1cm},clip,width=0.95\linewidth,cfbox={green 2pt 2pt}]{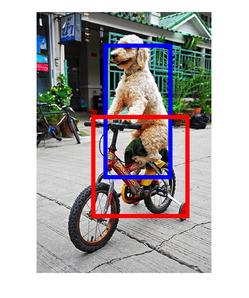}\\
       	\vspace{1.5ex}
    \end{minipage}
    \hspace{0.0025\textwidth}
    \begin{minipage}[t]{0.18\textwidth}
    	\centering
       	\includegraphics[trim={2.5cm 0.9cm 1.7cm 0.5cm},clip,width=0.95\linewidth,cfbox={green 2pt 2pt}]{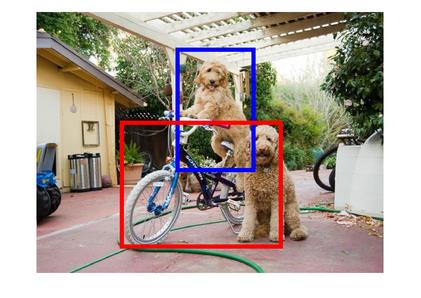}\\
       	\vspace{1.5ex}
    \end{minipage}
    \hspace{0.0025\textwidth}
    \begin{minipage}[t]{0.18\textwidth}
       \centering
       \includegraphics[trim={5.4cm 1cm 1.5cm 0.8cm},clip,width=0.95\linewidth,cfbox={green 2pt 2pt}]{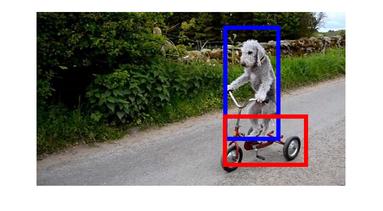}\\
       \vspace{1.5ex}
    \end{minipage}
    \hspace{0.0025\textwidth}
    \begin{minipage}[t]{0.18\textwidth}
    	\centering
       	\includegraphics[trim={-0.75cm 1cm -0.75cm 0.6cm},clip,width=0.95\linewidth,cfbox={red 2pt 2pt}]{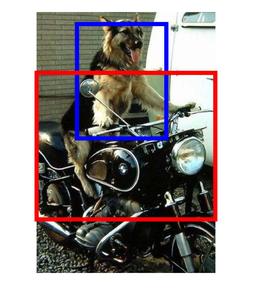}\\
      	\vspace{1.5ex}
    \end{minipage}
    
    \begin{minipage}[c]{0.24\textwidth}
    \vspace{-12ex}
    \small{
	(Q) \textbf{{\color{blue}building} {\textcolor{Green}{has}} {\color{red}wheel}} \\     	
	\vspace{-18pt}
	\par\noindent\rule{\textwidth}{0.4pt}
	(S) {\color{blue}truck} {\textcolor{Green}{has}} {\color{red}wheel} \\
	(S) {\color{blue}building} {\textcolor{Green}{has}} {\color{red}clock} \\
	(S) {\color{blue}bus} {\textcolor{Green}{has}} {\color{red}wheel} \\
	(S) {\color{blue}building} {\textcolor{Green}{has}} {\color{red}roof} \\
	(S) {\color{blue}cart} {\textcolor{Green}{has}} {\color{red}wheel} \\
	}
    \end{minipage}  
    \begin{minipage}[t]{0.18\textwidth}
    	\centering
       	\includegraphics[trim={1.3cm 1cm 2.2cm 0.5cm},clip,width=0.95\linewidth,cfbox={green 2pt 2pt}]{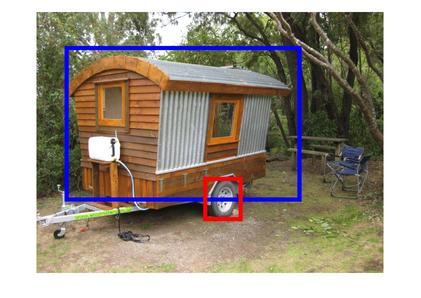}\\
       	\vspace{1.5ex}
    \end{minipage}
    \hspace{0.0025\textwidth}
    \begin{minipage}[t]{0.18\textwidth}
    	\centering
       	\includegraphics[trim={1.3cm 1.2cm 1.3cm 0.4cm},clip,width=0.95\linewidth,cfbox={green 2pt 2pt}]{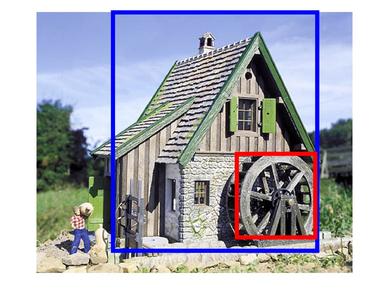}\\
       	\vspace{1.5ex}
    \end{minipage}
    \hspace{0.0025\textwidth}
    \begin{minipage}[t]{0.18\textwidth}
       \centering
       \includegraphics[trim={0.9cm 0.9cm 0.9cm 0.4cm},clip,width=0.95\linewidth,cfbox={green 2pt 2pt}]{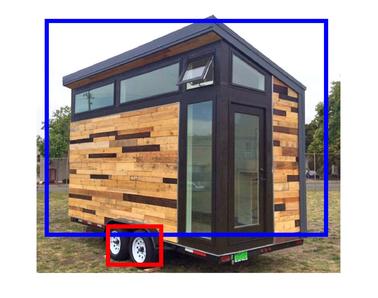}\\
       \vspace{1.5ex}
    \end{minipage}
    \hspace{0.0025\textwidth}
    \begin{minipage}[t]{0.18\textwidth}
    	\centering
       	\includegraphics[trim={1.1cm 0.9cm 2.4cm 0.6cm},clip,width=0.95\linewidth,cfbox={red 2pt 2pt}]{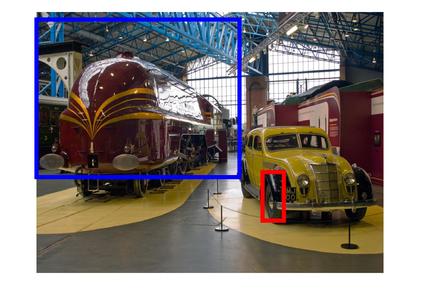}\\
      	\vspace{1.5ex}
    \end{minipage}

    \begin{minipage}[c]{0.24\textwidth}
    \vspace{-10ex}
    \small{
	(Q) \textbf{{\color{blue}person} {\textcolor{Green}{ride}} {\color{red}train}} \\     	
	\vspace{-18pt}
	\par\noindent\rule{\textwidth}{0.4pt}
	(S) {\color{blue}person} {\textcolor{Green}{ride}} {\color{red}boat} \\
	(S) {\color{blue}person} {\textcolor{Green}{ride}} {\color{red}horse} \\
	(S) {\color{blue}person} {\textcolor{Green}{ride}} {\color{red}motorcycle} \\
	(S) {\color{blue}person} {\textcolor{Green}{ride}} {\color{red}skateboard} \\
	(S) {\color{blue}person} {\textcolor{Green}{ride}} {\color{red}bike} \\
	}
    \end{minipage}  
    \begin{minipage}[t]{0.18\textwidth}
    	\centering
       	\includegraphics[trim={1.3cm 0.8cm 1.3cm 0.5cm},clip,width=0.95\linewidth,cfbox={green 2pt 2pt}]{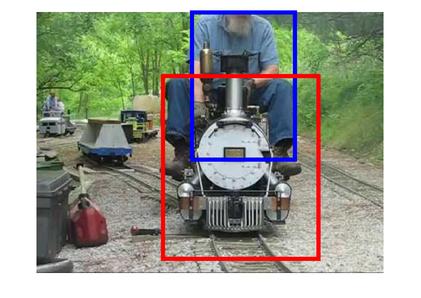}\\
       	\vspace{1.5ex}
    \end{minipage}
    \hspace{0.0025\textwidth}
    \begin{minipage}[t]{0.18\textwidth}
    	\centering
       	\includegraphics[trim={1.3cm 0.8cm 1.2cm 0.4cm},clip,width=0.95\linewidth,cfbox={green 2pt 2pt}]{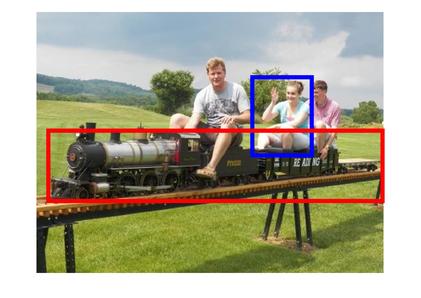}\\
       	\vspace{1.5ex}
    \end{minipage}
    \hspace{0.0025\textwidth}
    \begin{minipage}[t]{0.18\textwidth}
       \centering
       \includegraphics[trim={3.7cm 0.8cm 2.5cm 0.5cm},clip,width=0.95\linewidth,cfbox={green 2pt 2pt}]{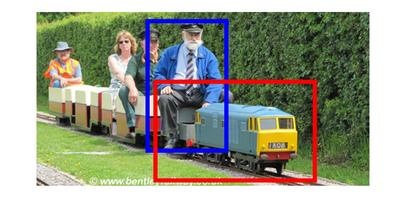}\\
       \vspace{1.5ex}
    \end{minipage}
    \hspace{0.0025\textwidth}
    \begin{minipage}[t]{0.18\textwidth}
    	\centering
       	\includegraphics[trim={1.8cm 0.9cm 1.2cm 0.5cm},clip,width=0.95\linewidth,cfbox={red 2pt 2pt}]{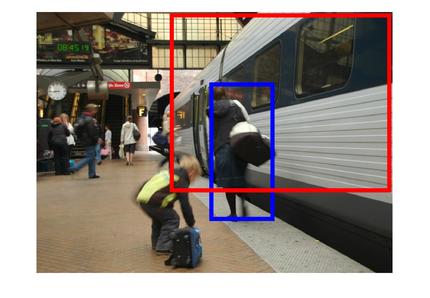}\\
      	\vspace{1.5ex}
    \end{minipage}

    \begin{minipage}[c]{0.24\textwidth}
    \vspace{-6ex}
    \small{
	(Q) \textbf{{\color{blue}car} {\textcolor{Green}{in}} {\color{red}building}} \\     	
	\vspace{-18pt}
	\par\noindent\rule{\textwidth}{0.4pt}
	(S) {\color{blue}car} {\textcolor{Green}{in}} {\color{red}street} \\
	(S) {\color{blue}bus} {\textcolor{Green}{in}} {\color{red}street} \\
	(S) {\color{blue}person} {\textcolor{Green}{in}} {\color{red}street} \\
	(S) {\color{blue}person} {\textcolor{Green}{in}} {\color{red}truck} \\
	(S) {\color{blue}person} {\textcolor{Green}{in}} {\color{red}bus} \\
	}
    \end{minipage}  
    \begin{minipage}[t]{0.18\textwidth}
    	\centering
       	\includegraphics[trim={1.1cm 0.9cm 2.2cm 0.5cm},clip,width=0.95\linewidth,cfbox={green 2pt 2pt}]{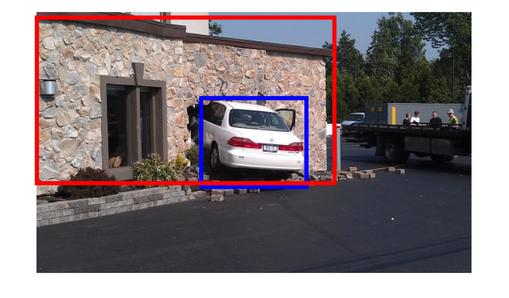}\\
       	\vspace{1.5ex}
    \end{minipage}
    \hspace{0.0025\textwidth}
    \begin{minipage}[t]{0.18\textwidth}
    	\centering
       	\includegraphics[trim={1.3cm 0.8cm 2.2cm 0.4cm},clip,width=0.95\linewidth,cfbox={green 2pt 2pt}]{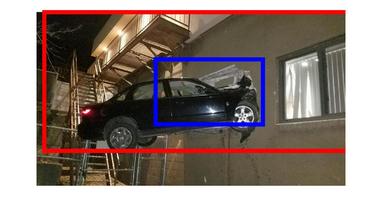}\\
       	\vspace{1.5ex}
    \end{minipage}
    \hspace{0.0025\textwidth}
    \begin{minipage}[t]{0.18\textwidth}
       \centering
       \includegraphics[trim={1.5cm 1.9cm 1.2cm 0.5cm},clip,width=0.95\linewidth,cfbox={green 2pt 2pt}]{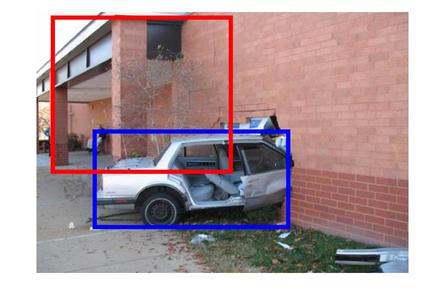}\\
       \vspace{1.5ex}
    \end{minipage}
    \hspace{0.0025\textwidth}
    \begin{minipage}[t]{0.18\textwidth}
    	\centering
       	\includegraphics[trim={1.2cm 0.9cm 2.5cm 0.4cm},clip,width=0.95\linewidth,cfbox={red 2pt 2pt}]{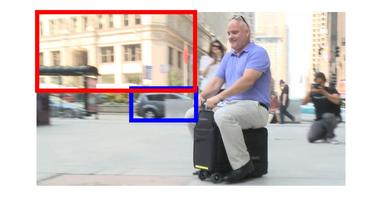}\\
      	\vspace{1.5ex}
    \end{minipage}

    \begin{minipage}[c]{0.24\textwidth}
    \vspace{-12ex}
    \small{
	(Q) \textbf{{\color{blue}cone} {\textcolor{Green}{on the top of}} {\color{red}person}} \\     	
	\vspace{-18pt}
	\par\noindent\rule{\textwidth}{0.4pt}
	(S) {\color{blue}tower} {\textcolor{Green}{on the top of}} {\color{red}building} \\
	(S) {\color{blue}roof} {\textcolor{Green}{on the top of}} {\color{red}building} \\
	(S) {\color{blue}laptop} {\textcolor{Green}{on the top of}} {\color{red}table} \\
	(S) {\color{blue}sky} {\textcolor{Green}{over}} {\color{red}person} \\
	(S) {\color{blue}umbrella} {\textcolor{Green}{over}} {\color{red}person} \\
	}
    \end{minipage}  
    \begin{minipage}[t]{0.18\textwidth}
    	\centering
       	\includegraphics[trim={2cm 0.8cm 2.9cm 0.5cm},clip,width=0.95\linewidth,cfbox={green 2pt 2pt}]{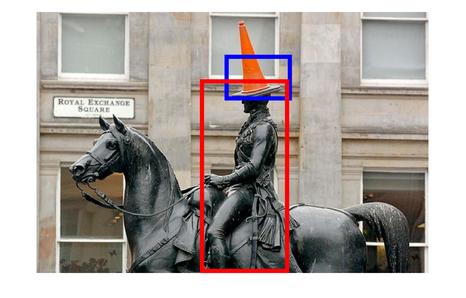}\\
       	\vspace{1.5ex}
    \end{minipage}
    \hspace{0.0025\textwidth}
    \begin{minipage}[t]{0.18\textwidth}
    	\centering
       	\includegraphics[trim={0cm 1.5cm 0cm 1.4cm},clip,width=0.95\linewidth,cfbox={green 2pt 2pt}]{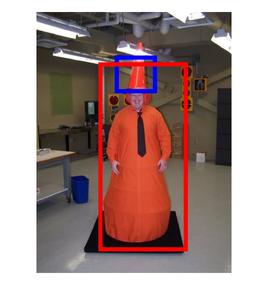}\\
       	\vspace{1.5ex}
    \end{minipage}
    \hspace{0.0025\textwidth}
    \begin{minipage}[t]{0.18\textwidth}
       \centering
       \includegraphics[trim={1.5cm 0.9cm 1.85cm 0.4cm},clip,width=0.95\linewidth,cfbox={green 2pt 2pt}]{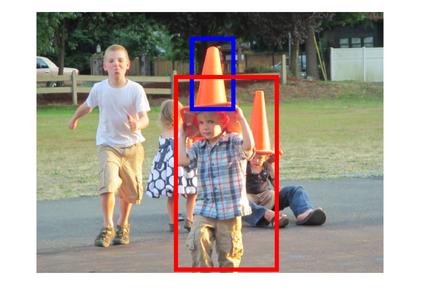}\\
       \vspace{1.5ex}
    \end{minipage}
    \hspace{0.0025\textwidth}
    \begin{minipage}[t]{0.18\textwidth}
    	\centering
       	\includegraphics[trim={0cm 2cm 0cm 1.5cm},clip,width=0.95\linewidth,cfbox={red 2pt 2pt}]{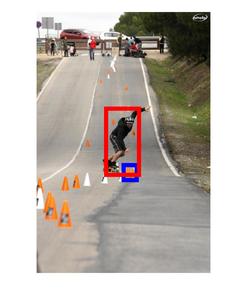}\\
      	\vspace{1.5ex}
    \end{minipage}

    \setlength\abovecaptionskip{10pt}
    \caption{\small {\bf Querying for unseen (unusual) triplets on the UnRel dataset.} Examples of retrieval using our model (p+vp+transfer). The query triplet is marked as (Q). The source triplets (S) seen in training are automatically selected by our model described in~\ref{part:model_2} and used to transfer the visual phrase embedding using the analogy transformation. 
  The automatically selected source triplets all appear relevant. Our method selects source triplets involving (1) a different subject (``dog ride bike" is transferred from ``person ride bike", ``building has wheel" is transferred from ``truck has wheel"), (2) different object (``person stand on horse" is transferred from ``person stand on sand"), or (3) different predicate (``cone on the top of person" is transferred from ``sky over person").}
    \label{fig:qualitative_unrel_supmat}
\end{figure*}

%% file: supmat_fig_cocoa.tex
\begin{figure*}[t]
\centering
	\begin{minipage}[t]{0.20\textwidth}
    \centering
    	\textit{Query (Q) / Source (S)}\\
    	\vspace{2ex}
	\end{minipage}	
	\hspace{0.01\textwidth}
	\begin{minipage}[t]{0.58\textwidth}
    	\centering
    \textit{	Top true positives}\\
    	\vspace{2ex}
	\end{minipage}
	\hspace{0.005\textwidth}
	\begin{minipage}[t]{0.18\textwidth}
    \centering
    \textit{	Top false positive}\\
    	\vspace{2ex}
	\end{minipage}

    \begin{minipage}[c]{0.24\textwidth}
    \vspace{-12ex}
    \small{
	(Q) \textbf{{\color{blue}person} {\textcolor{Green}{caress}} {\color{red}sheep}} \\     	
	\vspace{-18pt}
	\par\noindent\rule{\textwidth}{0.4pt}
	(S) {\color{blue}person} {\textcolor{Green}{herd}} {\color{red}sheep} \\
	(S) {\color{blue}person} {\textcolor{Green}{shear}} {\color{red}sheep} \\
	(S) {\color{blue}person} {\textcolor{Green}{walk}} {\color{red}sheep} \\
	(S) {\color{blue}person} {\textcolor{Green}{feed}} {\color{red}sheep} \\
	(S) {\color{blue}person} {\textcolor{Green}{herd}} {\color{red}cow} \\
	}
    \end{minipage}  
    \begin{minipage}[t]{0.18\textwidth}
    	\centering
       	\includegraphics[trim={1cm 0.9cm 1cm 0.6cm},clip,width=0.95\linewidth,cfbox={green 2pt 2pt}]{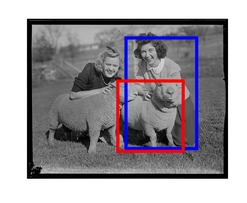}\\
       	\vspace{1.5ex}
    \end{minipage}
    \hspace{0.0025\textwidth}
    \begin{minipage}[t]{0.18\textwidth}
    	\centering
       	\includegraphics[trim={1.95cm 0.6cm 0.45cm 1cm},clip,width=0.95\linewidth,cfbox={green 2pt 2pt}]{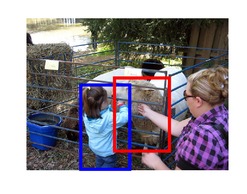}\\
       	\vspace{1.5ex}
    \end{minipage}
    \hspace{0.0025\textwidth}
    \begin{minipage}[t]{0.18\textwidth}
       \centering
       \includegraphics[trim={1.95cm 0.6cm 0.45cm 1cm},clip,width=0.95\linewidth,cfbox={green 2pt 2pt}]{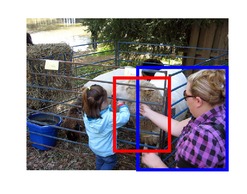}\\
       \vspace{1.5ex}
    \end{minipage}
    \hspace{0.0025\textwidth}
    \begin{minipage}[t]{0.18\textwidth}
    	\centering
       	\includegraphics[trim={0.4cm 0.95cm 0.3cm 1.5cm},clip,width=0.95\linewidth,cfbox={red 2pt 2pt}]{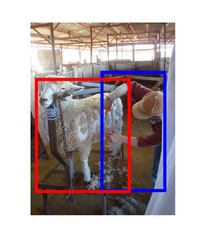}\\
      	\vspace{1.5ex}
    \end{minipage}

    \begin{minipage}[c]{0.24\textwidth}
    \vspace{-10ex}
    \small{
	(Q) \textbf{{\color{blue}person} {\textcolor{Green}{use}} {\color{red}laptop}} \\     	
	\vspace{-18pt}
	\par\noindent\rule{\textwidth}{0.4pt}
	(S) {\color{blue}person} {\textcolor{Green}{type on}} {\color{red}laptop} \\
	(S) {\color{blue}person} {\textcolor{Green}{read}} {\color{red}laptop} \\
	(S) {\color{blue}person} {\textcolor{Green}{type on}} {\color{red}keyboard} \\
	(S) {\color{blue}person} {\textcolor{Green}{text on}} {\color{red}cell phone} \\
	(S) {\color{blue}person} {\textcolor{Green}{control}} {\color{red}tv} \\
	}
    \end{minipage}  
    \begin{minipage}[t]{0.18\textwidth}
    	\centering
       	\includegraphics[trim={0.7cm 0.6cm 0.5cm 0.5cm},clip,width=0.95\linewidth,cfbox={green 2pt 2pt}]{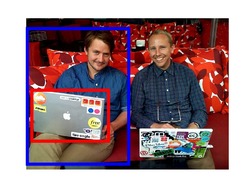}\\
       	\vspace{1.5ex}
    \end{minipage}
    \hspace{0.0025\textwidth}
    \begin{minipage}[t]{0.18\textwidth}
    	\centering
       	\includegraphics[trim={0.85cm 0.55cm 0.4cm 0.6cm},clip,width=0.95\linewidth,cfbox={green 2pt 2pt}]{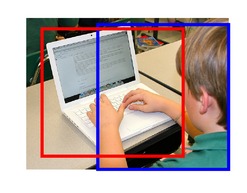}\\
       	\vspace{1.5ex}
    \end{minipage}
    \hspace{0.0025\textwidth}
    \begin{minipage}[t]{0.18\textwidth}
       \centering
       \includegraphics[trim={0.5cm 0.7cm 0.3cm 1.48cm},clip,width=0.95\linewidth,cfbox={green 2pt 2pt}]{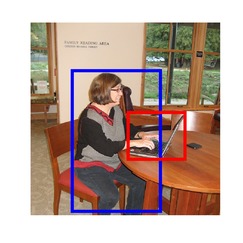}\\
       \vspace{1.5ex}
    \end{minipage}
    \hspace{0.0025\textwidth}
    \begin{minipage}[t]{0.18\textwidth}
    	\centering
       	\includegraphics[trim={-0.2cm 0.65cm -0.2cm 1.5cm},clip,width=0.95\linewidth,cfbox={red 2pt 2pt}]{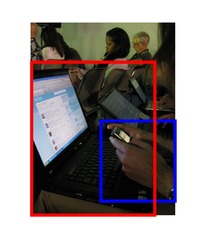}\\
      	\vspace{1.5ex}
    \end{minipage} 
    
    \begin{minipage}[c]{0.24\textwidth}
    \vspace{-12ex}
    \small{
	(Q) \textbf{{\color{blue}person} {\textcolor{Green}{touch}} {\color{red}horse}} \\     	
	\vspace{-18pt}
	\par\noindent\rule{\textwidth}{0.4pt}
	(S) {\color{blue}person} {\textcolor{Green}{hug}} {\color{red}horse} \\
	(S) {\color{blue}person} {\textcolor{Green}{pet}} {\color{red}horse} \\
	(S) {\color{blue}person} {\textcolor{Green}{kiss}} {\color{red}horse} \\
	(S) {\color{blue}person} {\textcolor{Green}{feed}} {\color{red}horse} \\
	(S) {\color{blue}person} {\textcolor{Green}{walk}} {\color{red}horse} \\
	}
    \end{minipage}  
    \begin{minipage}[t]{0.18\textwidth}
    	\centering
       	\includegraphics[trim={-0.3cm 0.7cm -0.4cm 1cm},clip,width=0.95\linewidth,cfbox={green 2pt 2pt}]{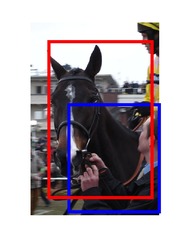}\\
       	\vspace{1.5ex}
    \end{minipage}
    \hspace{0.0025\textwidth}
    \begin{minipage}[t]{0.18\textwidth}
    	\centering
       	\includegraphics[trim={1cm 0.8cm 1.2cm 0.7cm},clip,width=0.95\linewidth,cfbox={green 2pt 2pt}]{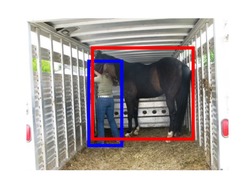}\\
       	\vspace{1.5ex}
    \end{minipage}
    \hspace{0.0025\textwidth}
    \begin{minipage}[t]{0.18\textwidth}
       \centering
       \includegraphics[trim={1cm 0.6cm 0.7cm 0.5cm},clip,width=0.95\linewidth,cfbox={green 2pt 2pt}]{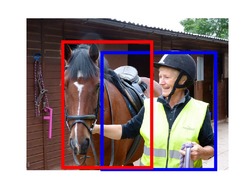}\\
       \vspace{1.5ex}
    \end{minipage}
    \hspace{0.0025\textwidth}
    \begin{minipage}[t]{0.18\textwidth}
    	\centering
       	\includegraphics[trim={0.8cm 0.3cm 0.4cm 0.15cm},clip,width=0.95\linewidth,cfbox={red 2pt 2pt}]{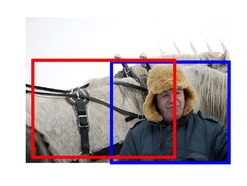}\\
      	\vspace{1.5ex}
    \end{minipage} 
        
    \begin{minipage}[c]{0.24\textwidth}
    \vspace{-16ex}
    \small{
	(Q) \textbf{{\color{blue}person} {\textcolor{Green}{get}} {\color{red}frisbee}} \\     	
	\vspace{-18pt}
	\par\noindent\rule{\textwidth}{0.4pt}
	(S) {\color{blue}person} {\textcolor{Green}{block}} {\color{red}frisbee} \\
	(S) {\color{blue}person} {\textcolor{Green}{throw}} {\color{red}frisbee} \\
	(S) {\color{blue}person} {\textcolor{Green}{catch}} {\color{red}frisbee} \\
	(S) {\color{blue}person} {\textcolor{Green}{hit}} {\color{red}sports ball} \\
	(S) {\color{blue}person} {\textcolor{Green}{block}} {\color{red}sports ball} \\
	}
    \end{minipage}  
    \begin{minipage}[t]{0.18\textwidth}
    	\centering
       	\includegraphics[trim={2.17cm 0.8cm 0.8cm 0.5cm},clip,width=0.95\linewidth,cfbox={green 2pt 2pt}]{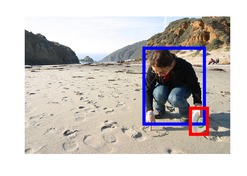}\\
       	\vspace{1.5ex}
    \end{minipage}
    \hspace{0.0025\textwidth}
    \begin{minipage}[t]{0.18\textwidth}
    	\centering
       	\includegraphics[trim={.10cm 0.9cm 0cm 1cm},clip,width=0.95\linewidth,cfbox={green 2pt 2pt}]{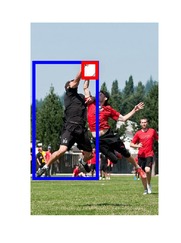}\\
       	\vspace{1.5ex}
    \end{minipage}
    \hspace{0.0025\textwidth}
    \begin{minipage}[t]{0.18\textwidth}
       \centering
       \includegraphics[trim={.10cm 0.9cm 0cm 1cm},clip,width=0.95\linewidth,cfbox={green 2pt 2pt}]{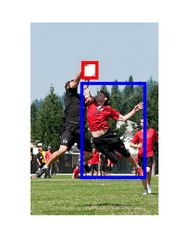}\\
       \vspace{1.5ex}
    \end{minipage}
    \hspace{0.0025\textwidth}
    \begin{minipage}[t]{0.18\textwidth}
    	\centering
       	\includegraphics[trim={0.025cm 0.8cm 0cm 1cm},clip,width=0.95\linewidth,cfbox={red 2pt 2pt}]{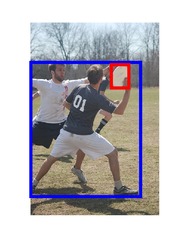}\\
      	\vspace{1.5ex}
    \end{minipage}     
    
    \begin{minipage}[c]{0.24\textwidth}
    \vspace{-12ex}
    \small{
	(Q) \textbf{{\color{blue}person} {\textcolor{Green}{chew}} {\color{red}toothbrush}} \\     	
	\vspace{-18pt}
	\par\noindent\rule{\textwidth}{0.4pt}
	(S) {\color{blue}person} {\textcolor{Green}{eat}} {\color{red}banana} \\
	(S) {\color{blue}person} {\textcolor{Green}{cook}} {\color{red}hot dog} \\
	(S) {\color{blue}person} {\textcolor{Green}{make}} {\color{red}hot dog} \\
	(S) {\color{blue}person} {\textcolor{Green}{inspect}} {\color{red}bottle} \\
	(S) {\color{blue}person} {\textcolor{Green}{eat}} {\color{red}apple} \\
	}
    \end{minipage}  
    \begin{minipage}[t]{0.18\textwidth}
    	\centering
       	\includegraphics[trim={-.50cm 0.7cm -.50cm 0.9cm},clip,width=0.95\linewidth,cfbox={green 2pt 2pt}]{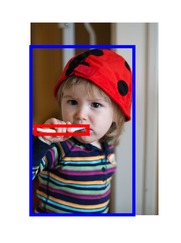}\\
       	\vspace{1.5ex}
    \end{minipage}
    \hspace{0.0025\textwidth}
    \begin{minipage}[t]{0.18\textwidth}
    	\centering
       	\includegraphics[trim={0.8cm 0.6cm 0.8cm 0.5cm},clip,width=0.95\linewidth,cfbox={green 2pt 2pt}]{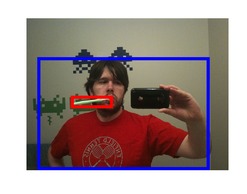}\\
       	\vspace{1.5ex}
    \end{minipage}
    \hspace{0.0025\textwidth}
    \begin{minipage}[t]{0.18\textwidth}
       \centering
       \includegraphics[trim={0.8cm 0.6cm 0.8cm 0.5cm},clip,width=0.95\linewidth,cfbox={green 2pt 2pt}]{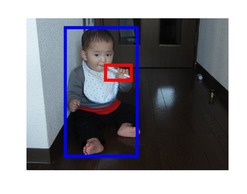}\\
       \vspace{1.5ex}
    \end{minipage}
    \hspace{0.0025\textwidth}
    \begin{minipage}[t]{0.18\textwidth}
    	\centering
       	\includegraphics[trim={0.025cm 1cm 0cm 1.1cm},clip,width=0.95\linewidth,cfbox={red 2pt 2pt}]{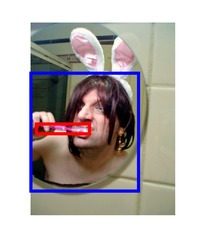}\\
      	\vspace{1.5ex}
    \end{minipage}

    \begin{minipage}[c]{0.24\textwidth}
    \vspace{-10ex}
    \small{
	(Q) \textbf{{\color{blue}person} {\textcolor{Green}{prepare}} {\color{red}kite}} \\     	
	\vspace{-18pt}
	\par\noindent\rule{\textwidth}{0.4pt}
	(S) {\color{blue}person} {\textcolor{Green}{fly}} {\color{red}kite} \\
	(S) {\color{blue}person} {\textcolor{Green}{launch}} {\color{red}kite} \\
	(S) {\color{blue}person} {\textcolor{Green}{inspect}} {\color{red}kite} \\
	(S) {\color{blue}person} {\textcolor{Green}{pull}} {\color{red}kite} \\
	(S) {\color{blue}person} {\textcolor{Green}{lie on}} {\color{red}surfboard} \\
	}
    \end{minipage}  
    \begin{minipage}[t]{0.18\textwidth}
    	\centering
       	\includegraphics[trim={1cm 0.55cm 0.8cm 0.5cm},clip,width=0.95\linewidth,cfbox={green 2pt 2pt}]{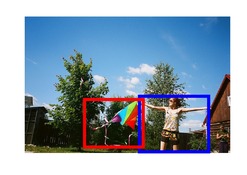}\\
       	\vspace{1.5ex}
    \end{minipage}
    \hspace{0.0025\textwidth}
    \begin{minipage}[t]{0.18\textwidth}
    	\centering
       	\includegraphics[trim={0.8cm 0.7cm 1cm 0.8cm},clip,width=0.95\linewidth,cfbox={green 2pt 2pt}]{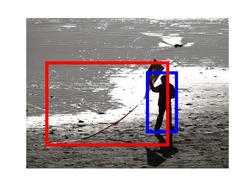}\\
       	\vspace{1.5ex}
    \end{minipage}
    \hspace{0.0025\textwidth}
    \begin{minipage}[t]{0.18\textwidth}
       \centering
       \includegraphics[trim={0.8cm 0.6cm 0.6cm 0.65cm},clip,width=0.95\linewidth,cfbox={green 2pt 2pt}]{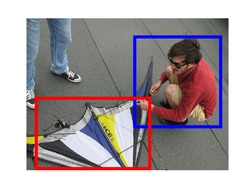}\\
       \vspace{1.5ex}
    \end{minipage}
    \hspace{0.0025\textwidth}
    \begin{minipage}[t]{0.18\textwidth}
    	\centering
       	\includegraphics[trim={0.9cm 0.6cm 1.05cm 0.6cm},clip,width=0.95\linewidth,cfbox={red 2pt 2pt}]{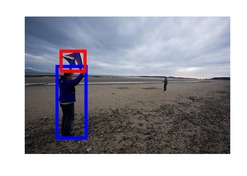}\\
      	\vspace{1.5ex}
    \end{minipage}  
    
    \setlength\abovecaptionskip{10pt}
    \caption{\small {\bf Querying for unseen (out of vocabulary) triplets on the COCO-a dataset.} Examples of retrieval using our model (s+o+vp+transfer). The query triplet is marked as (Q). The source triplets (S) seen in training are automatically selected by our model described in~\ref{part:model_2} and used to transfer the visual phrase embedding using the analogy transformation. 
  The automatically selected source triplets all appear relevant despite the difficulty that all predicates involved in the shown triplet queries are unseen at training time. The transfer to unseen predicates is made possible by the use of pre-trained word2vec embeddings. Given out-of-vocabulary triplets such as ``person use laptop" (row 2) or ``person touch horse" (row 3), our model automatically samples source triplets involving a relevant predicate such as ``person type on laptop" or ``person hug horse". However, we also observe that sometimes the out-of-vocabulary predicate is ambiguous (e.g. ``prepare" or ``get"), which makes it challenging to identify relevant source triplets among the set of available training triplets (e.g. ``person launch kite", ``person catch frisbee").}
    \label{fig:qualitative_cocoa_supmat}
    \vspace*{-5mm}
\end{figure*}

%% file: egpaper_for_review.bbl
\begin{thebibliography}{10}\itemsep=-1pt

\bibitem{Atzmon16}
Yuval Atzmon, Jonathan Berant, Vahid Kezami, Amir Globerson, and Gal Chechik.
\newblock Learning to generalize to new compositions in image understanding.
\newblock {\em arXiv:1608.07639}, 2016.

\bibitem{Aytar11}
Yusuf Aytar and Andrew Zisserman.
\newblock Tabula rasa: Model transfer for object category detection.
\newblock In {\em ICCV}, 2011.

\bibitem{Bansal2017}
Trapit Bansal, Arvind Neelakantan, and Andrew McCallum.
\newblock Relnet: End-to-end modeling of entities \& relations.
\newblock {\em arXiv:1706.07179}, 2017.

\bibitem{Battaglia2016}
Peter~W. Battaglia, Razvan Pascanu, Matthew Lai, Danilo~Jimeneze Rezende, and
  Koray Kavukcuoglu.
\newblock Interaction networks for learning about objects, relations and
  physics.
\newblock In {\em NIPS}, 2016.

\bibitem{Chao18}
Yu-Wei Chao, Yunfan Liu, Xieyang Liu, Huayi Zeng, and Jia Deng.
\newblock Learning to detect human-object interactions.
\newblock In {\em WACV}, 2018.

\bibitem{Chao15}
Yu-Wei Chao, Zhan Wang, Yugeng He, Jiaxuan Wang, and Jia Deng.
\newblock Hico: A benchmark for recognizing human-object interactions in
  images.
\newblock In {\em ICCV}, 2015.

\bibitem{Dai17}
Bo Dai, Yuqi Zhang, and Dahua Lin.
\newblock Detecting visual relationships with deep relational networks.
\newblock In {\em CVPR}, 2017.

\bibitem{Divvala2014}
Santosh~Kumar Divvala, Ali Farhadi, and Carlos Guestrin.
\newblock Learning everything about anything: Webly-supervised visual concept
  learning.
\newblock In {\em CVPR}, 2014.

\bibitem{iCAN}
Chen Gao, Yuliang Zou, and Jia-Bin Huang.
\newblock Ican: Instance-centric attention network for human-object interaction
  detection.
\newblock In {\em BMVC}, 2018.

\bibitem{Detectron2018}
Ross Girshick, Ilija Radosavovic, Georgia Gkioxari, Piotr Doll\'{a}r, and
  Kaiming He.
\newblock Detectron.
\newblock \url{https://github.com/facebookresearch/detectron}, 2018.

\bibitem{Gkioxari18}
Georgia Gkioxari, Ross Girshick, and Kaiming He.
\newblock Detecting and recognizing human-object interactions.
\newblock In {\em CVPR}, 2018.

\bibitem{Gupta15}
Saurabh Gupta and Jitendra Malik.
\newblock Visual role semantic labeling.
\newblock {\em arXiv:1505.04474}, 2015.

\bibitem{Hwang18}
Seong~Jae Hwang, Sathya~N. Ravi, Zirui Tao, Hyunwoo~J. Kim, Maxwell~D. Collins,
  and Vikas Singh.
\newblock Tensorize, factorize and regularize: Robust visual relationship
  learning.
\newblock In {\em CVPR}, 2018.

\bibitem{Izadinia2015}
Hamid Izadinia, Fereshteh Sadeghi, Santosh~Kumar Divvala, Yejin Choi, and Ali
  Farhadi.
\newblock Segment-phrase table for semantic segmentation, visual entailment and
  paraphrasing.
\newblock In {\em ICCV}, 2015.

\bibitem{jenatton2012}
Rodolphe Jenatton, Nicolas~L Roux, Antoine Bordes, and Guillaume~R Obozinski.
\newblock A latent factor model for highly multi-relational data.
\newblock In {\em NIPS}, 2012.

\bibitem{Johnson2015}
Justin Johnson, Andrej Karpathy, and Li Fei-Fei.
\newblock Densecap: Fully convolutional localization networks for dense
  captioning.
\newblock In {\em CVPR}, 2016.

\bibitem{Johnson15a}
Justin Johnson, Ranjay Krishna, Michael Stark, Li-Jia Li, David~A Shamma,
  Michael~S Bernstein, and Li Fei-Fei.
\newblock Image retrieval using scene graphs.
\newblock In {\em CVPR}, 2015.

\bibitem{Karpathy2014}
Andrej Karpathy and Li Fei-Fei.
\newblock Deep visual-semantic alignments for generating image descriptions.
\newblock In {\em CVPR}, 2015.

\bibitem{Karpathy2014a}
Andrej Karpathy, Armand Joulin, and Li Fei-Fei.
\newblock Deep fragment embeddings for bidirectional image sentence mapping.
\newblock In {\em NIPS}, 2014.

\bibitem{Kato18}
Keizo Kato, Yi Li, and Abhinav Gupta.
\newblock Compositional learning for human object interaction.
\newblock In {\em ECCV}, 2018.

\bibitem{Adam}
Diederik~P. Kingma and Jimmy Ba.
\newblock Adam: A method for stochastic optimization.
\newblock In {\em ICLR}, 2015.

\bibitem{Kipf2016}
Thomas~N. Kipf and Max Welling.
\newblock Semi-supervised classification with graph convolutional networks.
\newblock In {\em ICLR}, 2016.

\bibitem{Krishna2016}
Ranjay Krishna, Yuke Zhu, Oliver Groth, Justin Johnson, Kenji Hata, Joshua
  Kravitz, Stephanie Chen, Yannis Kalantidis, Li-Jia Li, David~A Shamma,
  Michael Bernstein, and Li Fei-Fei.
\newblock Visual genome: Connecting language and vision using crowdsourced
  dense image annotations.
\newblock In {\em IJCV}, 2016.

\bibitem{Yikang17bis}
Yikang Li, Wanli Ouyang, and Xiaogang Wang.
\newblock Vip-cnn: A visual phrase reasoning convolutional neural network for
  visual relationship detection.
\newblock In {\em CVPR}, 2017.

\bibitem{Lin2017FeaturePN}
Tsung-Yi Lin, Piotr Doll{\'a}r, Ross~B. Girshick, Kaiming He, Bharath
  Hariharan, and Serge~J. Belongie.
\newblock Feature pyramid networks for object detection.
\newblock In {\em CVPR}, 2017.

\bibitem{Lin2014a}
Tsung-Yi Lin, Michael Maire, Serge Belongie, James Hays, Pietro Perona, Deva
  Ramanan, Piotr Doll{\'a}r, and C~Lawrence Zitnick.
\newblock Microsoft {COCO}: Common objects in context.
\newblock In {\em ECCV}, 2014.

\bibitem{Lu16}
Cewu Lu, Ranjay Krishna, Michael Bernstein, and Li Fei-Fei.
\newblock Visual relationship detection with language priors.
\newblock In {\em ECCV}, 2016.

\bibitem{mikolov2013distributed}
Tomas Mikolov, Ilya Sutskever, Kai Chen, Greg~S Corrado, and Jeff Dean.
\newblock Distributed representations of words and phrases and their
  compositionality.
\newblock In {\em NIPS}, 2013.

\bibitem{Misra17}
Ishan Misra, Abhinav Gupta, and Martial Hebert.
\newblock From red wine to red tomato: Composition with context.
\newblock In {\em CVPR}, 2017.

\bibitem{Peyre17}
Julia Peyre, Ivan Laptev, Cordelia Schmid, and Josef Sivic.
\newblock Weakly-supervised learning of visual relations.
\newblock In {\em ICCV}, 2017.

\bibitem{plummerPLCLC2016}
Bryan~A. Plummer, Arun Mallya, Christopher~M. Cervantes, Julia Hockenmaier, and
  Svetlana Lazebnik.
\newblock Phrase localization and visual relationship detection with
  comprehensive linguistic cues.
\newblock In {\em ICCV}, 2017.

\bibitem{Plummer15}
Bryan~A. Plummer, Liwei Wang, Chris~M. Cervantes, Juan~C. Caicedo, Julia
  Hockenmaier, and Svetlana Lazebnik.
\newblock Flickr30k entities: Collecting region-to-phrase correspondences for
  richer image-to-sentence models.
\newblock In {\em ICCV}, 2015.

\bibitem{Qi18}
Siyuan Qi, Wenguan Wang, Baoxiong Jia, Jianbing Shen, and Song-Chun Zhu.
\newblock Learning human-object interactions by graph parsing neural networks.
\newblock In {\em ECCV}, 2018.

\bibitem{ramanathan15}
Vignesh Ramanathan, Congcong Li, Jia Deng, Wei Han, Zhen Li, Kunlong Gu, Yang
  Song, Samy Bengio, Chuck Rossenberg, and Li Fei-Fei.
\newblock Learning semantic relationships for better action retrieval in
  images.
\newblock In {\em CVPR}, 2015.

\bibitem{Reed2015a}
Scott~E. Reed, Yi Zhang, Yuting Zhang, and Honglak Lee.
\newblock Deep visual analogy-making.
\newblock In {\em NIPS}, 2015.

\bibitem{ren15}
Shaoqing Ren, Kaiming He, Ross Girshick, and Jian Sun.
\newblock Faster {R-CNN}: Towards real-time object detection with region
  proposal networks.
\newblock In {\em NIPS}, 2015.

\bibitem{Ronchi2015}
Matteo~Ruggero Ronchi and Pietro Perona.
\newblock Describing common human visual actions in images.
\newblock In {\em BMVC}, 2015.

\bibitem{sadeghi2015viske}
Fereshteh Sadeghi, Santosh~K Divvala, and Ali Farhadi.
\newblock Viske: Visual knowledge extraction and question answering by visual
  verification of relation phrases.
\newblock In {\em CVPR}, 2015.

\bibitem{Sadeghi2015}
Fereshteh Sadeghi, C.~Lawrence Zitnick, and Ali Farhadi.
\newblock Visalogy: Answering visual analogy questions.
\newblock In {\em NIPS}, 2015.

\bibitem{Sadeghi2011}
Mohammad~Amin Sadeghi and Ali Farhadi.
\newblock Recognition using visual phrases.
\newblock In {\em CVPR}, 2011.

\bibitem{Santoro17}
Adam Santoro, David Raposo, David~G.T. Barrett, Mateusz Malinowski, Razvan
  Pascanu, Peter Battaglia, and Timothy Lillicrap.
\newblock A simple neural network module for relational reasoning.
\newblock In {\em NIPS}, 2017.

\bibitem{Shen18}
Liyue Shen, Serena Yeung, Judy Hoffman, Greg Mori, and Li Fei-Fei.
\newblock Scaling human-object interaction recognition through zero-shot
  learning.
\newblock In {\em WACV}, 2018.

\bibitem{tsne}
Laurens Van~der Maaten and Geoffrey Hinton.
\newblock Visualizing data using t-sne.
\newblock {\em JMLR}, 2008.

\bibitem{Wang16}
Liwei Wang, Yin Li, and Svetlana Lazebnik.
\newblock Learning deep structure-preserving image-text embeddings.
\newblock In {\em CVPR}, 2016.

\bibitem{Yu17}
Ruichi Yu, Ang Li, Vlad~I. Morariu, and Larry~S. Davis.
\newblock Visual relationship detection with internal and external linguistic
  knowledge distillation.
\newblock In {\em ICCV}, 2017.

\bibitem{Zhang17}
Hanwang Zhang, Zawlin Kyaw, Shih-Fu Chang, and Tat-Seng Chua.
\newblock Visual translation embedding network for visual relation detection.
\newblock In {\em CVPR}, 2017.

\bibitem{Ji18}
Ji Zhang, Yannis Kalantidis, Marcus Rohrbach, Manohar Paluri, Ahmed Elgammal,
  and Mohamed Elhoseiny.
\newblock Large-scale visual relationship understanding.
\newblock In {\em AAAI}, 2019.

\bibitem{Bohan17}
Bohan Zhuang, Lingqiao Liu, Chunhua Shen, and Ian Reid.
\newblock Towards context-aware interaction recognition for visual relationship
  detection.
\newblock In {\em ICCV}, 2017.

\end{thebibliography}
